\documentclass{article}

\usepackage{arxiv}

\usepackage[utf8]{inputenc} 
\usepackage[T1]{fontenc}    
\usepackage{hyperref}       
\usepackage{url}            
\usepackage{booktabs}       
\usepackage{amsfonts}       
\usepackage{nicefrac}       
\usepackage{microtype}      
\usepackage{lipsum}		
\usepackage{graphicx}
\usepackage{natbib}
\usepackage{doi}
\usepackage{caption}
\usepackage{subcaption}
\usepackage{amsmath, amssymb}
\usepackage[table]{xcolor}
\usepackage[warn]{textcomp}
\usepackage{epsfig}
\usepackage{verbatim}
\usepackage{multirow}
\usepackage{txfonts}

\title{Concurrent build direction, part segmentation, and topology optimization for additive manufacturing using neural networks}

\date{} 					

\author{Hongrui Chen \qquad Aditya Joglekar \qquad Kate S. Whitefoot \qquad \textbf{Levent Burak Kara}\thanks{Address all correspondences to lkara@cmu.edu} \\ Department of Mechanical Engineering\\Carnegie Mellon University\\ Pittsburgh, PA, 15213, USA}

\hypersetup{
pdftitle={Concurrent build direction, part segmentation, and topology optimization for additive manufacturing using neural networks},
pdfsubject={q-bio.NC, q-bio.QM},
pdfauthor={Hongrui Chen},
pdfkeywords={FNNTO_arxiv},
}

\begin{document}
\maketitle

\begin{abstract}
{

We propose a neural network-based approach to topology optimization that aims to reduce the use of support structures in additive manufacturing. Our approach uses a network architecture that allows the simultaneous determination of an optimized: (1) part segmentation, (2) the topology of each part, and (3) the build direction of each part that collectively minimize the amount of support structure. Through training, the network learns a material density and segment classification in the continuous 3D space. Given a problem domain with prescribed load and displacement boundary conditions, the neural network takes as input 3D coordinates of the voxelized domain as training samples and outputs a continuous density field. Since the neural network for topology optimization learns the density distribution field, analytical solutions to the density gradient can be obtained from the input-output relationship of the neural network. We demonstrate our approach on several compliance minimization problems with volume fraction constraints, where support volume minimization is added as an additional criterion to the objective function. We show that simultaneous optimization of part segmentation along with the topology and print angle optimization further reduces the support structure, compared to a combined print angle and topology optimization without segmentation.}
\end{abstract}

\section{INTRODUCTION}
Additive manufacturing has become a popular method of manufacturing topology optimized components. Depositing materials layer by layer, additive manufacturing can manufacture topology optimization designs that are challenging or impossible for traditional subtractive machining. However, additive manufacturing requires support structures for overhang regions, which are common in the resulting shapes of topology optimization designs. Support structures require additional materials that do not contribute to the final product while taking up extra time during fabrication. Moreover, extra effort is required to remove support material afterward. It is possible to reduce support material without modifying the topology by reorienting the print direction or segmenting the part into smaller components \cite{Nie2020}. However, this approach is limited to the extent that it can eliminate support material. Simultaneously optimizing build orientation and segmentation with the part’s topology could result in further elimination of support structures.

Previously, optimization for reducing support structures has been implemented in both Solid Isotropic Material with Penalization (SIMP), level-set method \cite{Qian2017,Mirzendehdel2016} and part consolidation \cite{Nie2020}. Wang and Qian further developed the SIMP based method by adding simultaneous build orientation optimization\cite{Wang2020}. While overhang surfaces occur at the boundary of geometry, Heaviside projection is required to push SIMP results closer to 0 and 1 for a clear boundary. Chandrasekhar and Suresh’s work on neural network-based topology optimization demonstrated the possibility of direct topology optimization using neural network \cite{Chandrasekhar2021}. A neural network is a universal function approximator. The SIMP density field can be directly represented by neural network activation functions with the additional benefit of a crisp and differentiable boundary. Nie et al. demonstrated that print orientation optimization for each sub-assembly obtained after part decomposition can provide a greater reduction in support structure compared to print direction optimization for the large part \cite{Nie2020}. Support structures may be further reduced by adding segmentation to the simultaneous print direction and overhang optimization. We develop upon the neural network approach used by Chandrasekhar and Suresh and extend Wang and Qian's method of overhang detection, to achieve concurrent build direction, part segmentation and topology optimization to reduce support structures. 

Our approach combines print direction, segmentation, and topology optimization through a modular framework using neural networks. Within this modular framework, topology and segmentation is handled through Radial Basis function Neural Networks (RBNN) whereas print direction is determined through a single layer perceptron which serves as an optimizer that fits within the modular neural network framework. This modular neural network architecture allows combined optimization by selectively turning on and off different modules. The RBNN learns a density field for which it receives 3D coordinates as input and outputs density values. From this input-output relationship, we can obtain an analytical solution to the density gradient which directly correlates to the surface gradient used for overhang minimization. Finally, through combined build direction, segmentation, and topology optimization, we achieve superior results compared to previous methods. 

The implementation of this work can be found at: https://github.com/HongRayChen/seg-angle-topopt\\

Our main contributions are:
\begin{itemize}
  \item   An analytical solution to the density gradient obtained from the input-output relationship of the neural network
  \item   Combined build direction, part segmentation, and topology optimization
  \item   A modular neural network architecture that allows combined optimization and selectively turning on and off different modules
\end{itemize}

\begin{figure*}
\centering





\includegraphics[width=\textwidth]{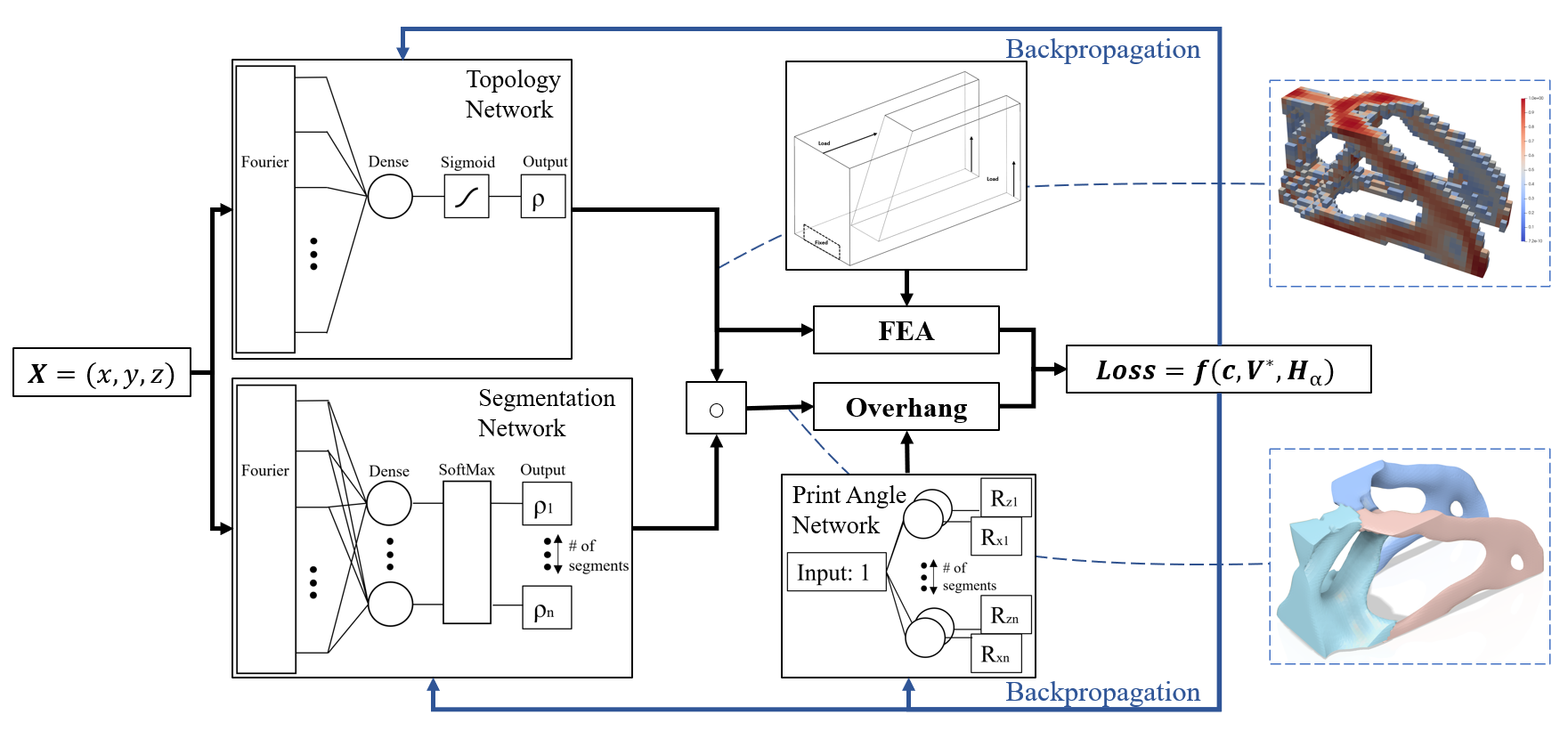}

\caption{Our topology, segmentation, and print angle neural network.  The topology and segmentation networks use a radial basis layer to transform  input coordinates into the Fourier space. The print angle network consists of a single layer of neurons that encode the print angle parameters, which is compatible with the overall neural network-based architecture. On the top right is the voxelized output of the topology network. On the bottom right is an upsampled and smoothed output with a unique color assigned to each segment.  }

\label{fig:comb_arch}
\end{figure*}
\section{RELATED WORK}

Our review focuses on studies that highlight additive manufacturing oriented topology optimization and integration of machine learning and topology optimization.

\textit{Support structure minimization}.  A large body of work has investigated overhang edge detection and penalization for support structure minimization. Thompson et al. \cite{Thompson2016} explored support structures and their relationship to additive manufacturing costs and opportunities. Brackett et al. \cite{Brackett2011} first explored an approach to penalize overhang edges through edge detection in 2D. Several studies inferred surface gradients through approximation based on near neighbors. Leary et al. \cite{Leary2014} created self-supporting structures with boundaries obtained through 4 or 8 element connectivity. Overhang edges can also be obtained by scanning elements below. Gaynor and Guest \cite{Gaynor2014} demonstrated the detection of three sets of overhang angles. Mhapsekar et al. \cite{Mhapsekar2018} also used neighbor detection with additional height penalty. Overhang regions can also be excluded from layer-wise filtering \cite{Langelaar2016}. Ven et al. \cite{van2018continuous} used front propagation to detect overhang surfaces. Zhang et al. \cite{Zhang2018} created self-supporting structures through polygon modifications. Support structure placement can be minimized while reducing the residual stress \cite{Cheng2019Support}. Most aforementioned methods lack accurate density gradient calculation due to the discretized grid of topology optimization. More emphasis on improving the accuracy of the density gradient calculation had been made through later research. Using the level set method, overhang control can be implemented by calculating the topological derivative \cite{Mirzendehdel2016}. Liu and To \cite{Liu2017Support} demonstrated using level set for toolpath simulation to consider both raster direction and build direction. Zhang et al. \cite{Zhang2019} fitted a linear field based on the topology generated and improved the density gradient calculation accuracy. Qian et al. \cite{Qian2017} used a PDE filter to obtain a more accurate density gradient. Through PDE filtering of density gradient calculation, self-supporting structures, boundary slope control, and print angle optimization have been added for simultaneous optimization with topology \cite{Mezzadri2018,Wang2019,Wang2020}. We base our overhang detection method on Wang and Qian’s work which integrates print angle from the vector dot product of the print angle with filtered density gradient with the distinction of (1) accurate, differentiable density gradient from the neural network which topology optimization is directly executed on and no filtering required, (2) additional height penalization which minimizes tall and slim support structures that increase the possibility of print failure.

\textit{Print direction and manufacturability optimization}. Print direction is the direction in which a part is oriented during additive manufacturing. Chandrasekhar et al. \cite{Chandrasekhar2020} demonstrated the importance of print direction on structure performance by optimizing the print direction of fiber-reinforced additive manufactured components. Ulu et al. \cite{Ulu2015} trained a surrogate model from running FEA in different directions for which an optimal build orientation can be determined from gradient descent on surrogate model. Ulu et al. \cite{Ulu2019} also demonstrated the advantages of concurrent optimization of the total production cost. By small adjustment to the shape and print direction optimization, models can also be better suited for additive manufacturing \cite{Ulu2019Mnfblty}. Nie et al. \cite{Nie2020} conducted optimization for minimum production cost considering part consolidation while demonstrating the tradeoff between no and full consolidation. Part layout, including rotation of each individual component, is optimized. Nie et al. demonstrated that for metal additive manufacturing, segmentation leads to a reduction in shadow volume. However, the topology of each component remains unchanged which motivates us to add part segmentation to our framework to further reduce shadow volume.

\textit{Topology optimization with machine learning}. Both data-driven and real-time approaches have been explored for the application of topology optimization. Data-driven topology optimization aims to learn a neural network model from a database of topology optimization results which speeds up the process. Surrogate neural network model has been used for microstructure design \cite{White2019}. Many other methods rely on Convolutional Neural Networks (CNN) for their capabilities to learn from a large set of image data. Banga et al. \cite{Banga2018} used a 3D encoder-decoder CNN to generate 3D topology results with a 4$\%$ reduction in computation time. Behzadi and Ilieş \cite{Behzadi2021} used deep transfer learning with CNN. Zheng et al. \cite{Zheng2021} used U-net CNN for 3D topology synthesis. Machine learning can also generate an initial guess for topology optimization to speed up the convergence \cite{Cang2019}. More accurate synthesis can be achieved from generative adversarial networks based on physical fields over the initial domain \cite{Nie2021}. U-Net was also used for improving the manufacturablity of designs for metal additive manufacturing \cite{iyer2021pato}. CNN demonstrated the capabilities of rapid topology synthesis but CNN alone cannot guarantee the mechanical performance of the result. A real-time data-driven hybrid approach trained the neural network during optimization to learned optimization sensitivities to further accelerate topology optimization \cite{Chi2021}. Chandrasekhar and Suresh \cite{Chandrasekhar2021} first explored a real-time approach where the neural network directly optimizes the density field of SIMP. It guaranteed the mechanical performance of topology optimization as the density field is parameterized by the weights of the neural network. Chandrasekhar and Suresh \cite{Chandrasekhar2021Fourier} also explored Fourier projection based neural network for length scale control. Application of multi-material topology optimization is also explored with a similar concept \cite{Chandrasekhar2021MM}. Multi-material topology optimization is achieved with a multi-layer perceptron with a softmax layer attached at the end which serves to assign the material to each segment. Chandrasekhar and Suresh’s \cite{Chandrasekhar2021MM} research demonstrated the benefit of a real-time neural network-based approach where boundaries can be differentiable. In our work, we further exploit this advantage for the application of minimizing overhang while leveraging the segmentation capabilities to partition the topology for print direction optimization of each individual segment.


\begin{figure*}
\centering

\begin{subfigure}[t]{0.3\textwidth}
\centering
\includegraphics[width=0.5\textwidth]{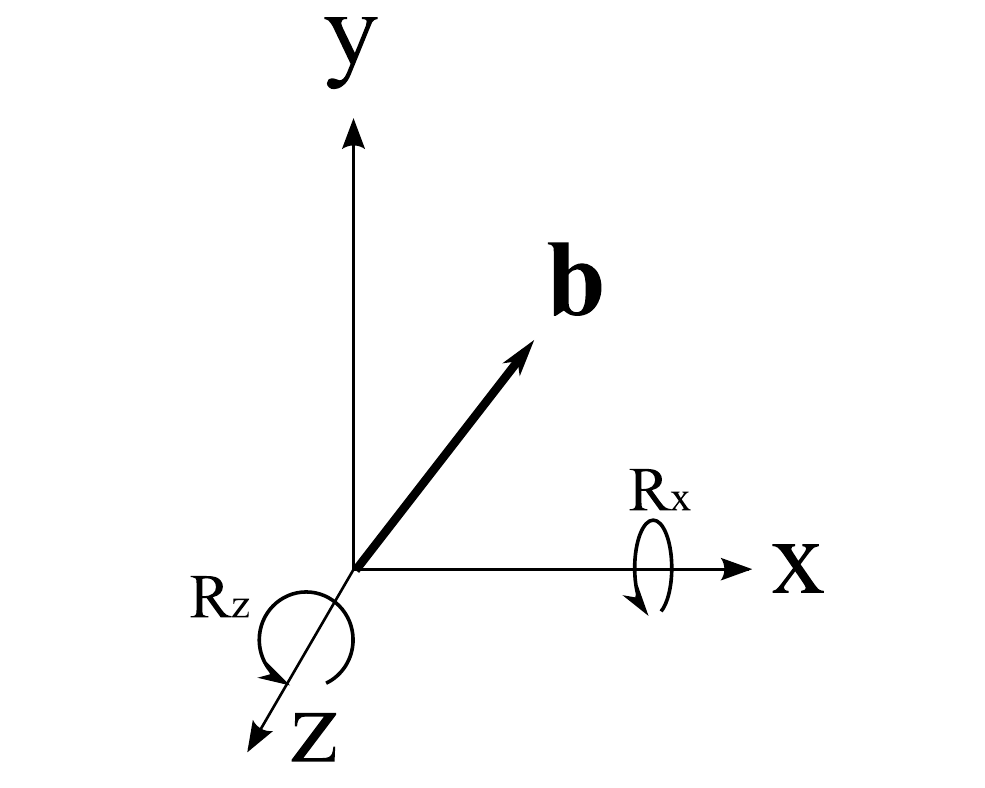}
\caption{Definition of the coordinate system}
\end{subfigure}
\qquad
\begin{subfigure}[t]{0.3\textwidth}
\centering
\includegraphics[width=\textwidth]{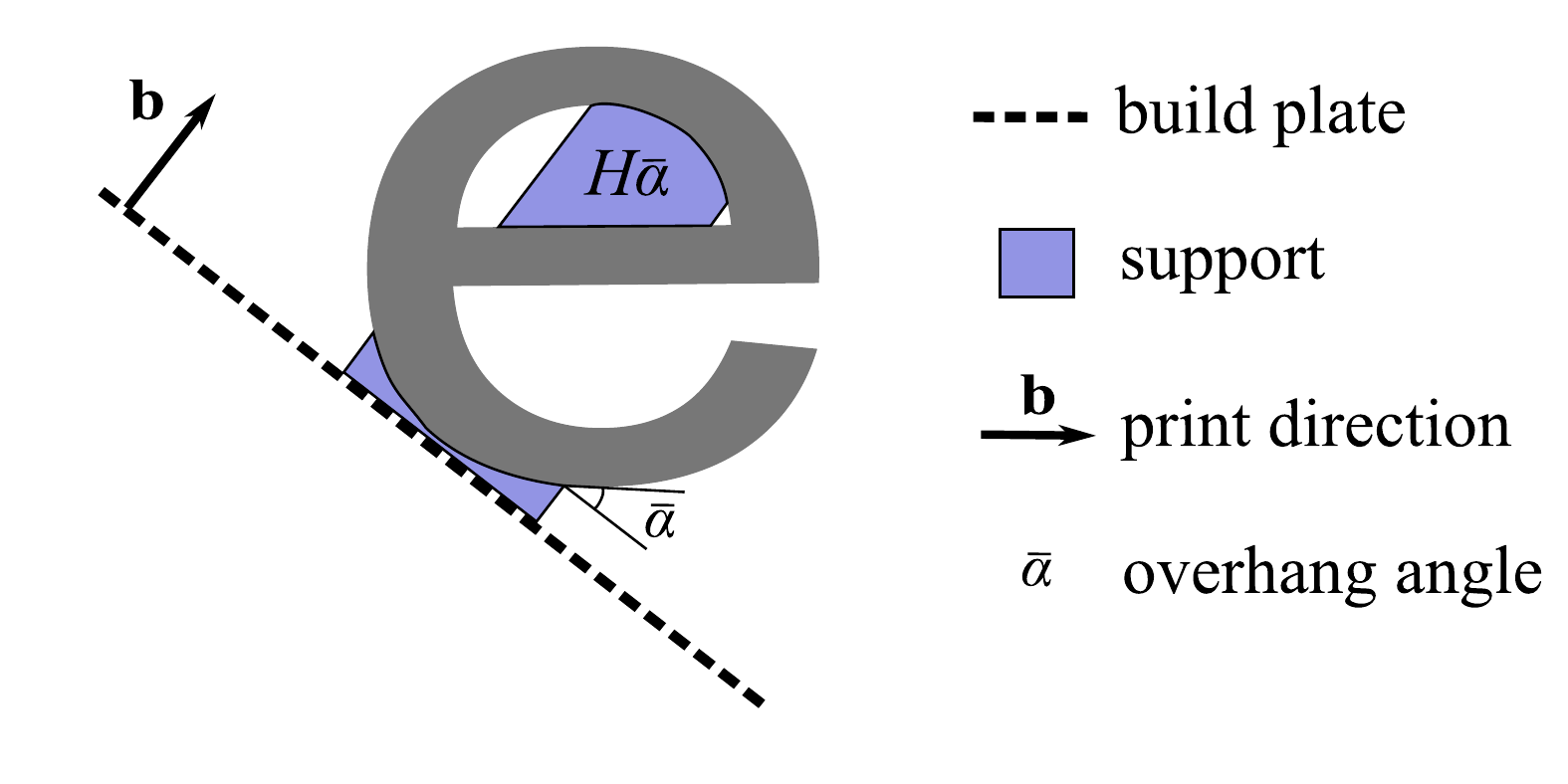}
\caption{Overhang and support definition for an `e' shaped structure}
\end{subfigure}
\qquad
\begin{subfigure}[t]{0.3\textwidth}
\centering
\includegraphics[width=\textwidth]{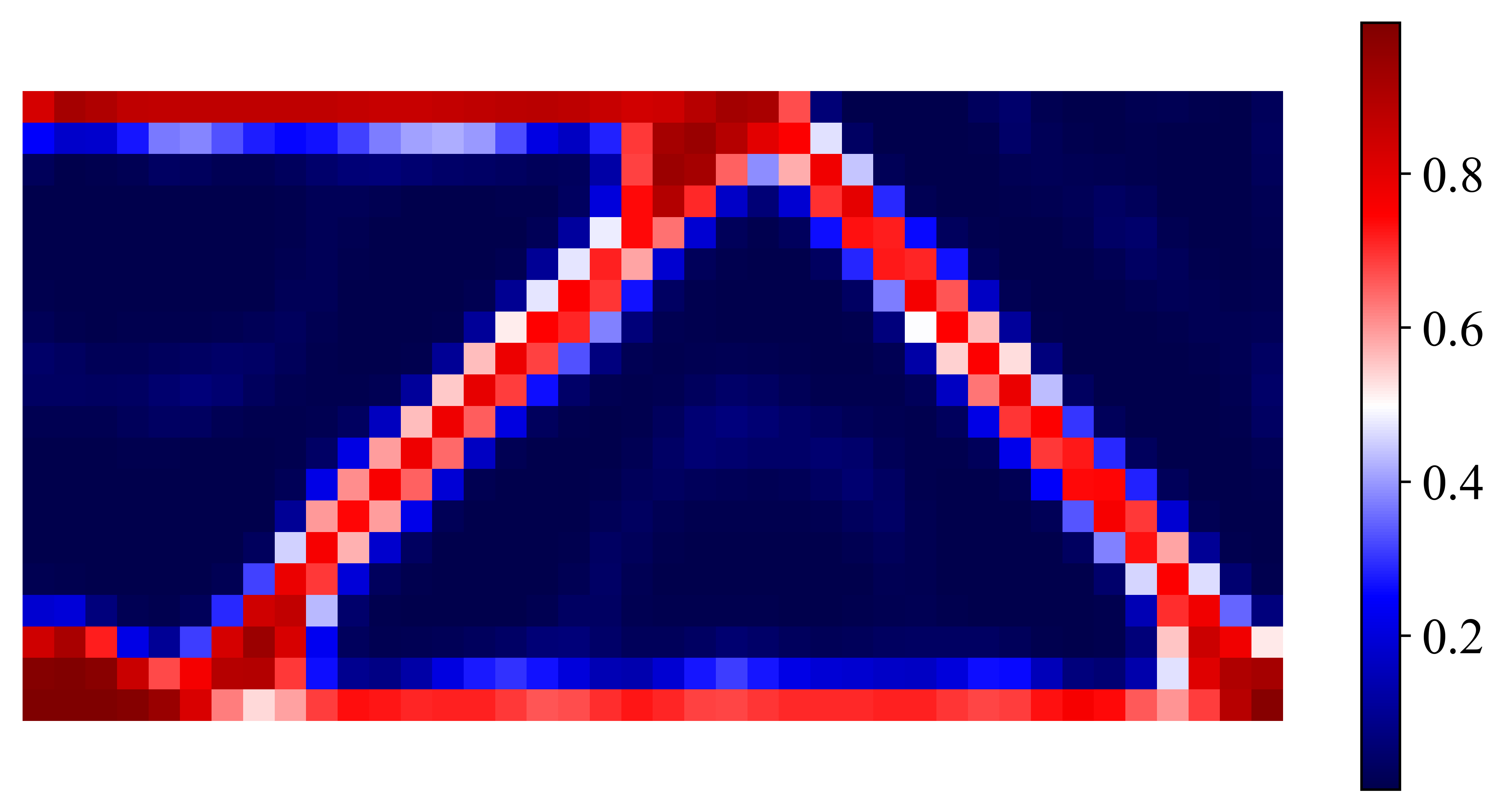}
\caption{$\rho$}
\end{subfigure}

\begin{subfigure}[t]{0.3\textwidth}
\centering
\includegraphics[width=\textwidth]{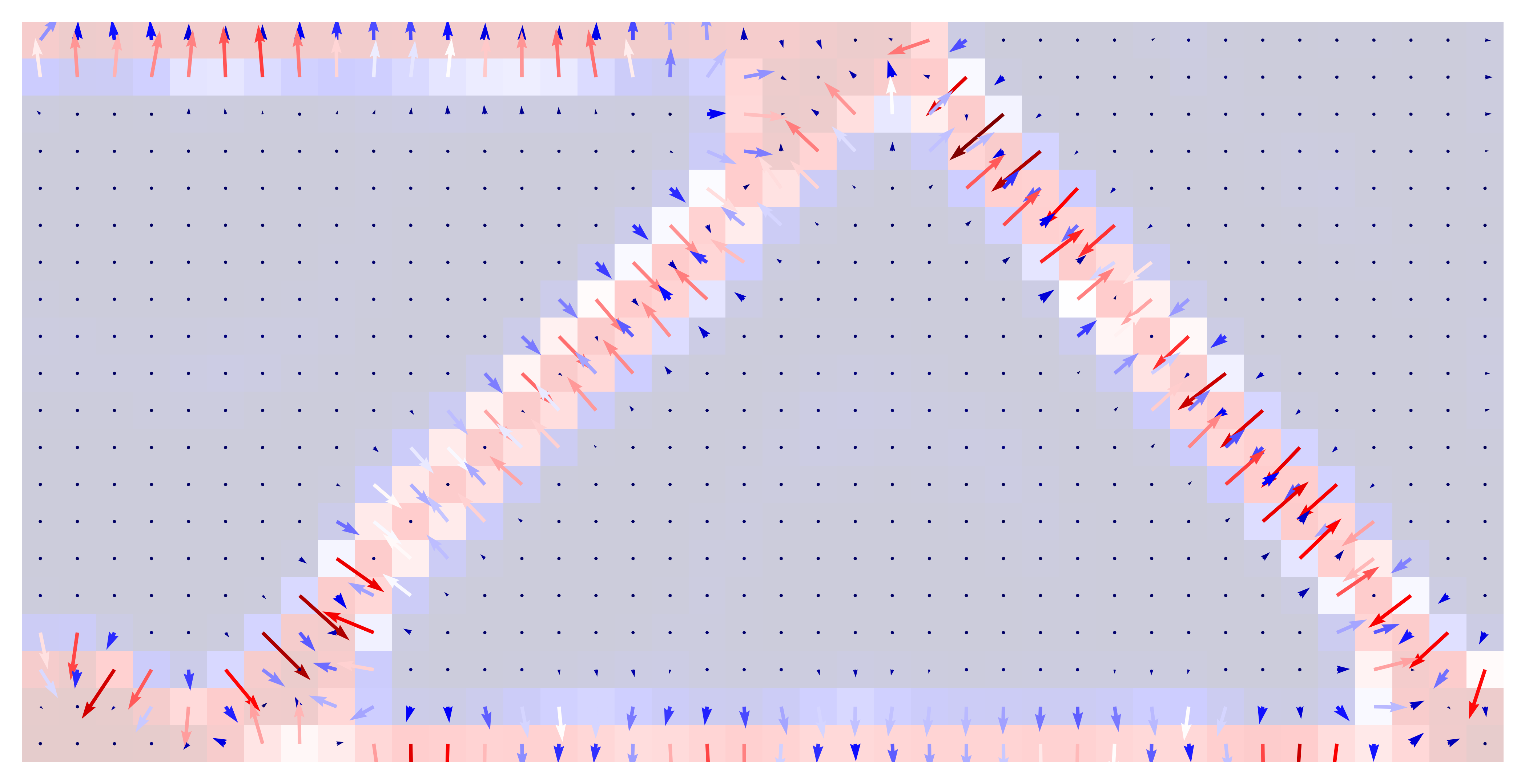}
\caption{$\nabla\rho$}
\end{subfigure}
\qquad
\begin{subfigure}[t]{0.45\textwidth}
\centering
\includegraphics[width=\textwidth]{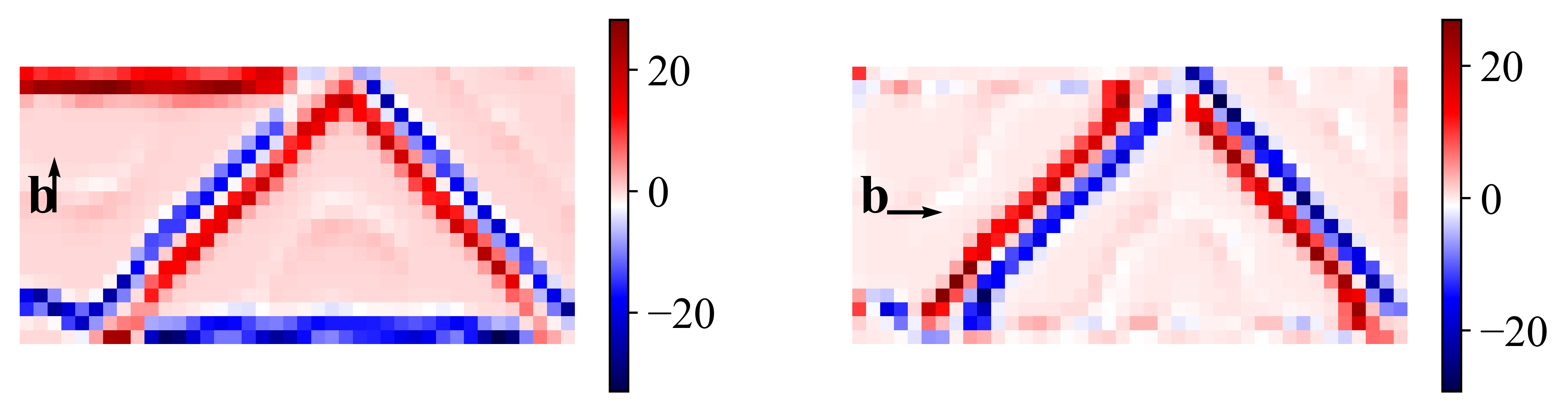}
\caption{$\textbf{b}\cdot \nabla\rho$}
\end{subfigure}

\begin{subfigure}[t]{0.45\textwidth}
\centering
\includegraphics[width=\textwidth]{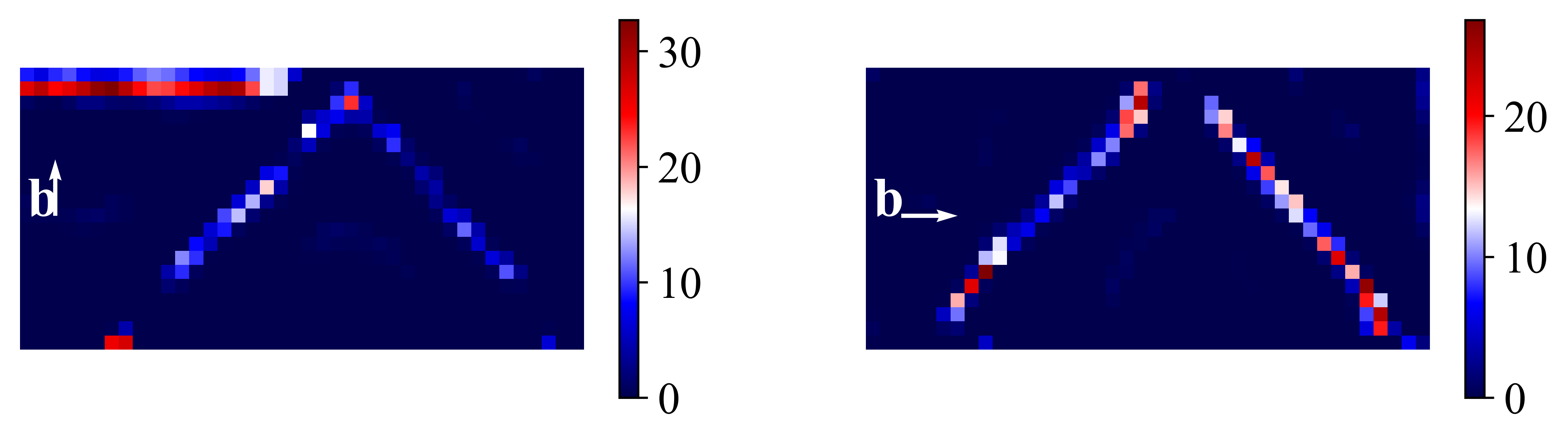}
\caption{$P_{\bar{\alpha}}$}
\end{subfigure}
\qquad
\begin{subfigure}[t]{0.45\textwidth}
\centering
\includegraphics[width=\textwidth]{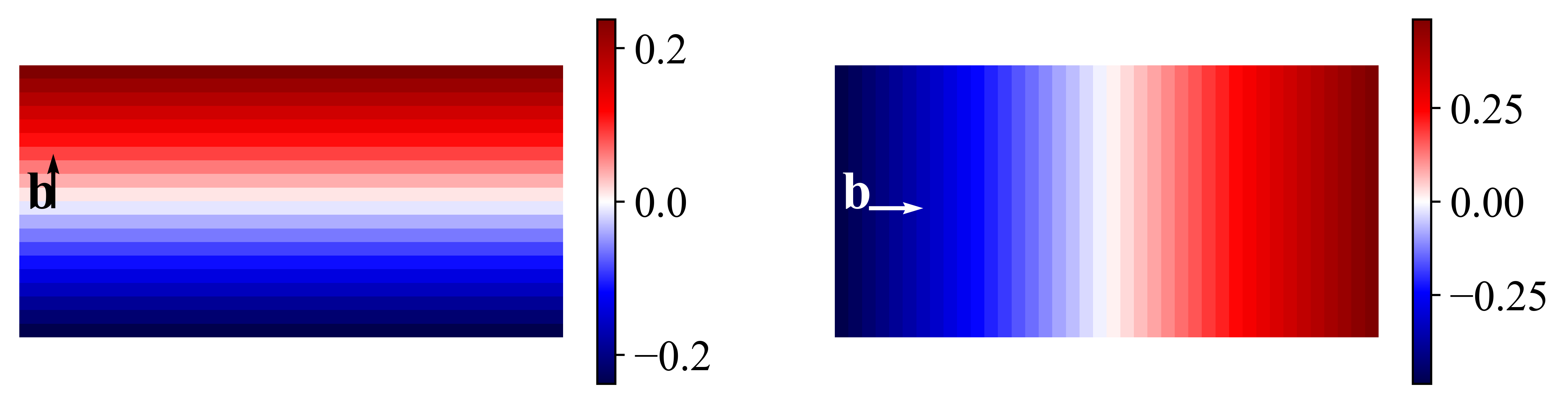}
\caption{Rotated Y-axis coordinates}
\end{subfigure}

\begin{subfigure}[t]{0.45\textwidth}
\centering
\includegraphics[width=\textwidth]{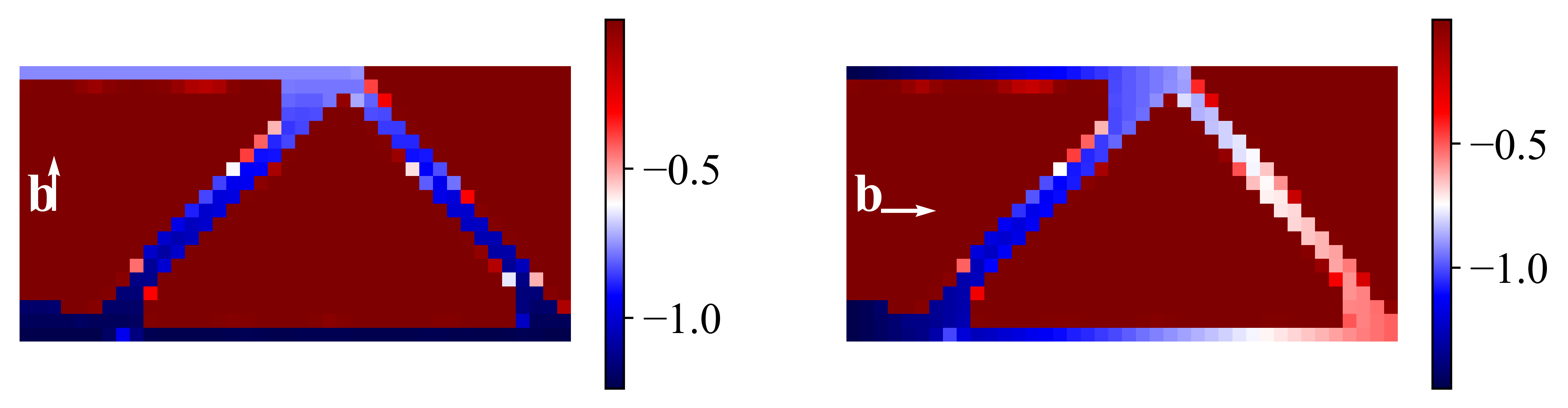}
\caption{$(\textbf{X}_y - 1.0)\tilde{\rho}$}
\end{subfigure}
\qquad
\begin{subfigure}[t]{0.45\textwidth}
\centering
\includegraphics[width=\textwidth]{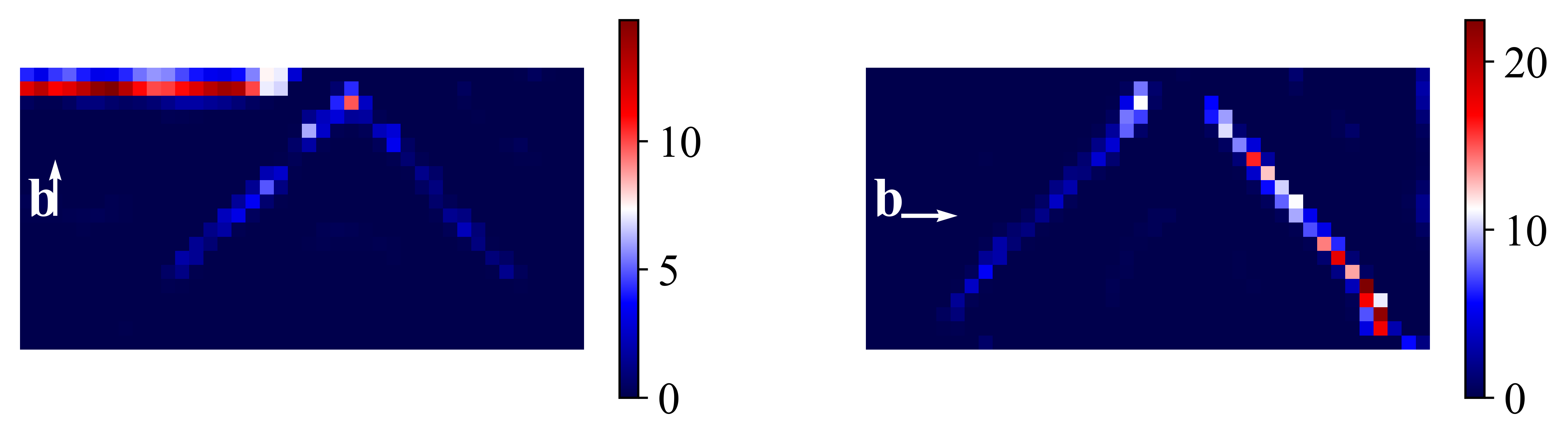}
\caption{$H_{\bar{\alpha}}$}
\end{subfigure}

\centering
\caption{Overhang regions $P_{\bar{\alpha}}$ can be detected based on Heaviside filtering of the print direction dot product with the density gradient $\textbf{b}\cdot \nabla\rho$. We further augment the overhang regions by multiplying these regions with their distance from the build plate to penalize overhang regions $H_{\bar{\alpha}}$ that are higher above the build plate which require more support material.}

\label{fig:sv_calc}
\end{figure*}

\section{PROPOSED METHOD}
The goal of this work is to reduce the support structure for topology optimization through segmentation and print angle optimization. We decompose the functionality of generating topology, segmentation, and print angle for each segment into three modular neural networks (Figure \ref{fig:comb_arch}). These three modular networks enable us to selectively turn on and off different modules depending on the complexity of the problem. Using only topology neural networks, we can run standalone topology optimization similar to SIMP.  While print angle optimization can be added for simultaneous print direction and topology optimization similar to Wang and Qian’s approach \cite{Wang2020}. With segmentation added, concurrent optimization can be achieved. These three modules are linked together by a combined loss function. Backpropagation of the loss function into each module is handled by the machine learning package TensorFlow \cite{Abadi2016}.

\subsection{Neural Network}
The topology network $T(\textbf{X})$ learns a density field which is different compared to typical topology optimization which represents the density field as a finite element mesh. The topology neural network takes in 3D coordinates $\textbf{X} = (x,y,z)$ as inputs and outputs the density value $\rho$ at each coordinate point. The 3D coordinates represent the center of each element in the design domain. For standalone topology optimization, a batch of 3D coordinates that correspond to a 3D mesh grid is fed into the topology network. The output is then sent to the Finite Element Analysis (FEA) solver. The solver outputs the compliance which is combined with volume fraction as a loss. The loss is then backpropagated to adjust the weight of each neuron of the topology network. 

For the topology network design, we employed a modified RBNN. The first layer of the neural network consists of a radial basis function layer. This layer applies a non-linear transformation of the 3D coordinates input. We use TensorFlow’s “RandomFourierFeatures” with uniformly sampled kernel to implement this layer \cite{Abadi2016}. For repeatability, we initialize the kernel location in a linear 3D grid. In contrast to the neural network architecture from Chandrasekhar and Suresh \cite{Chandrasekhar2021Fourier}, the second layer is a fully-connected layer with only one neuron. The final layer consists of a sigmoid activation layer. The sigmoid activation function guarantees the output is between 0 and 1. The main feature of this network design is the removal of a batch normalization layer. Together with the shift-invariant Gaussian kernel, we can choose a flexible batch size as input without worrying about the statistical distribution of the input. The flexible batch size input is useful when enforcing non-design regions. We can upsample the 3D coordinate input or only sample specific regions of the density field to manipulate the resolution of the discretized visualization. The topology network can be formulated as follows:
\begin{equation}
T(\textbf{X}) = \sigma(\cos(\textbf{X}\times \textbf{K} +  \textbf b_{kernel}) \times \textbf{W} + \textbf b_{dense} )
\end{equation}
Where $\textbf{X}$ is the 3D coordinate input, $\textbf{X}=(x,y,z)$. 
$\sigma$ is the Sigmoid activation function. 
$\textbf b_{kernel}$ is the Bias of the RBNN layer. 
$\textbf K$ are Kernels for the RBNN layer. 
$\textbf W$ are Weights for the dense layer.
$\textbf b_{dense}$ is Bias for the dense layer. 

The initial step for conducting overhang analysis is to obtain the density gradient $\nabla\rho$. Since the topology neural network learns a continuous density field with 3D coordinates as input, obtaining the $\nabla\rho$ is straightforward in this case. We simply use the automatic differentiation from TensorFlow to track the coordinate input and the density output to obtain $\frac{\delta \rho}{\delta \textbf{X} }$\cite{Abadi2016}. $\frac{\delta \rho}{\delta \textbf{X} }$ is the analytical density gradient at each 3D coordinate input. Mathematically, $\nabla\rho$ can be derived from using chain rule by taking the derivative w.r.t input $\textbf{X}$, $T'(\textbf{X}) = \frac{\delta \rho}{\delta \textbf{X}}$:

\begin{equation}
T'(\textbf{X}) = (-\sin(\textbf{X}\times \textbf K + \textbf b_{kernel})(\sigma'(\textbf M) \times \textbf W^{T})) \odot \textbf K^{T}
\end{equation}
Where $\textbf M = \cos(\textbf{X}\times \textbf K + \textbf b_{kernel}) \times \textbf W + \textbf b_{dense}$, $
\sigma'(\textbf M) = \sigma(\textbf M)(1-\sigma(\textbf M)$. $\odot$ represents broadcasting multiplication. 

Part segmentation refers to breaking up a larger part into smaller segments. Each segment is a monolithic part and will have a unique print angle. The geometry of each segment will be optimized by jointly taking into account overhang minimization. The segmentation network $S(\textbf{X})$ is different from the topology network in two ways. First, the segmentation network second layer has the same number of neurons as the number of segments that is requested. Second, the final layer is a Softmax layer to guarantee partition of unity. The segmentation network can be formulated as: 

\begin{equation}
S(\textbf{X}) = s(\cos(\textbf{X}\times \textbf K + 
\textbf b_{kernel}) \times \textbf W + \textbf b_{dense})
\end{equation}
Where $s(x)$ is the Softmax operator $s(z) = \frac{e^{z_i}}{\sum_{l=1}^n e^{z_l}}$. 
Using chain rule, the derivative of $S(\textbf{X})$ with respect to  input $\textbf{X}$ is:
\begin{equation}
S'(\textbf{X}) = (-\sin(\textbf{X}\times \textbf K + \textbf b_{kernel})({s_i}'(\textbf M) \times \textbf W^{T})) \odot \textbf K^{T} 
\end{equation}\\
Where $\textbf M = \cos(\textbf{X}\times \textbf K + \textbf b_{kernel}) \times \textbf W + \textbf b_{dense}$, ${s_i}'(\textbf M) = s_i(\textbf M)(\bold 1\{i=j\}-s_j(\textbf M))$;\, $i,j = 1,2,...n_{segs}$

Segmentation is applied by multiplying the topology network output with the segmentation network output. The topology segmentation multiplication is similar to a Hadamard product where we duplicate the output from the topology network number of segments times and multiply with segmentation network output (Figure \ref{fig:comb_arch}, "$\circ$" represents the Hadamard prodcut). We also initialize the segmentation with an inverse distance field. This serves as an initial guess for the segmentation while reducing the possibility of disconnected segments. 
\begin{equation}
\rho = T(\textbf{X})\circ S(\textbf{X})
\label{eq_overhang_def}
\end{equation}
The print angle for each segment is generated from the print angle network. The print angle network consists of only one layer of neurons which takes in 1 as a constant input. The neurons are directly connected to the output which means the weight of each neuron directly corresponds to the output. The purpose of designing it as a neural network is to allow the print angle to be integrated with the combined framework. In addition, by directly manipulating the weights, we can configure the initial condition of print angle for each segment. 


\begin{figure*}
\centering

\begin{subfigure}[b]{0.23\textwidth}
\centering
\includegraphics[width=\textwidth]{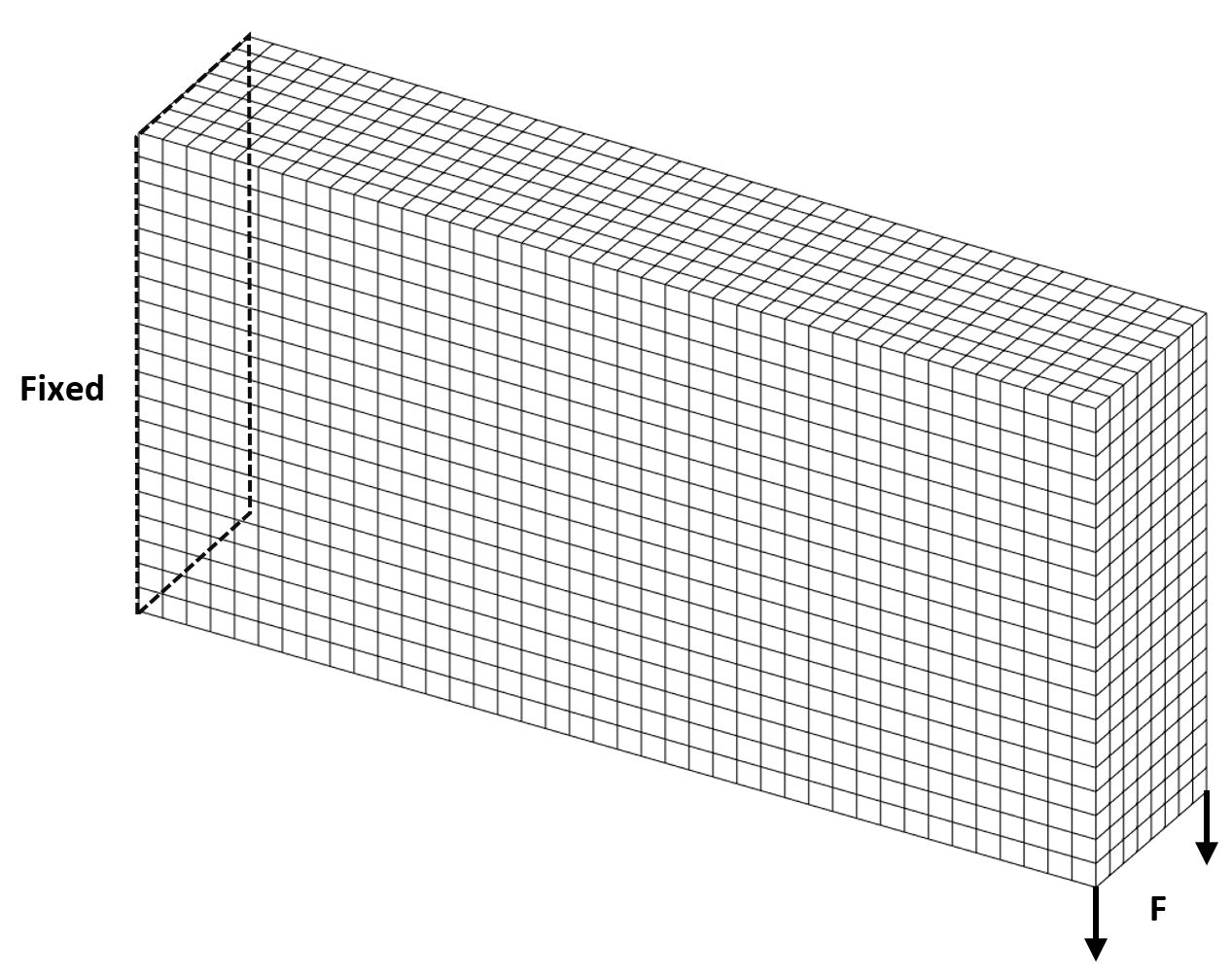}
\caption{Boundary condition}
\end{subfigure}
\hfill
\begin{subfigure}[b]{0.23\textwidth}
\centering
\includegraphics[width=\textwidth]{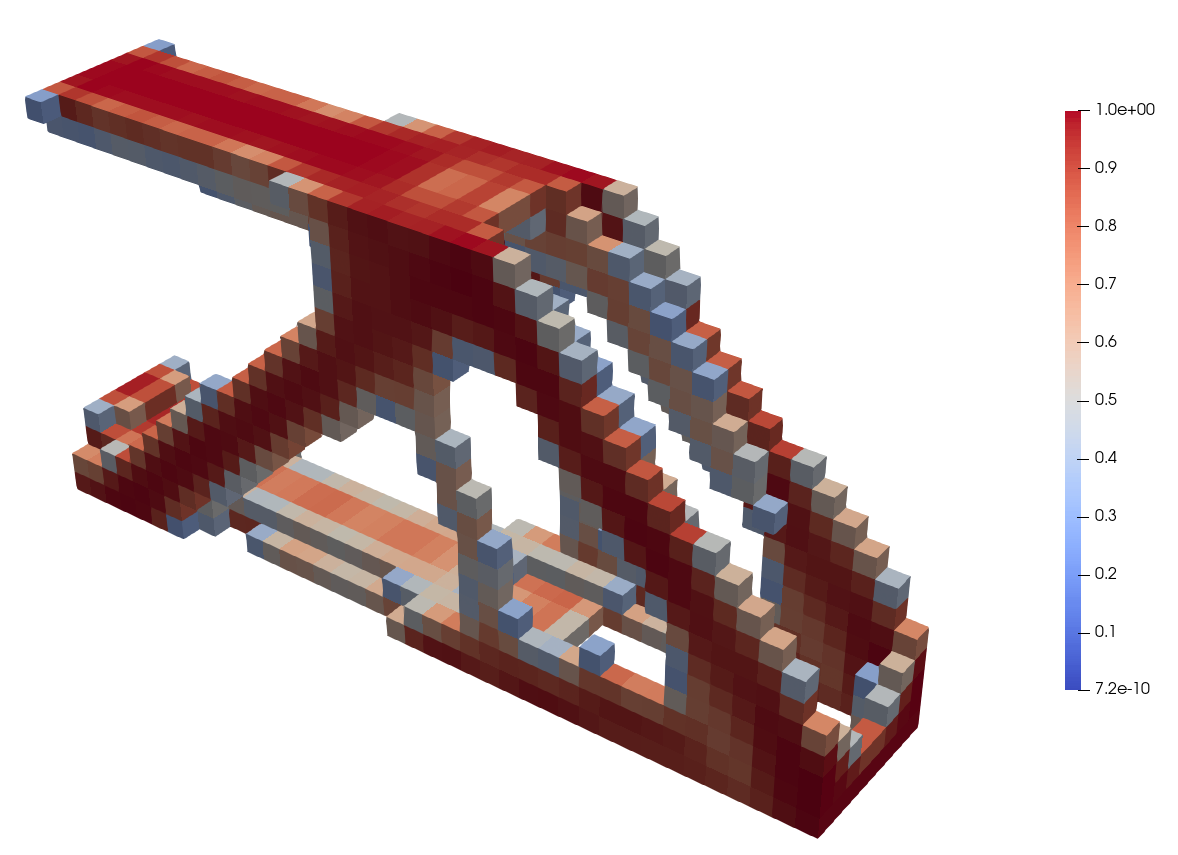}
\caption{$c=40.6$}
\end{subfigure}
\hfill
\begin{subfigure}[b]{0.23\textwidth}
\centering
\includegraphics[width=\textwidth]{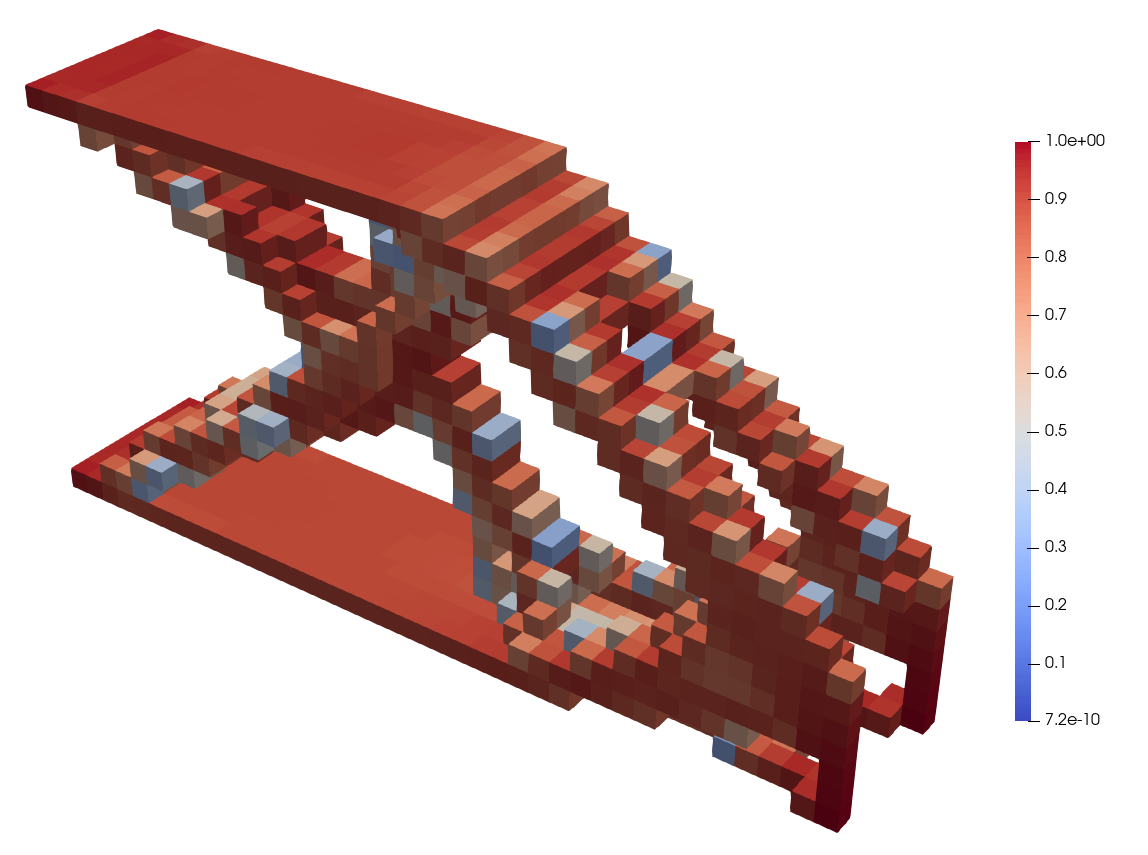}
\caption{$c=37.8$}
\end{subfigure}
\hfill
\begin{subfigure}[b]{0.23\textwidth}
\centering
\includegraphics[width=\textwidth]{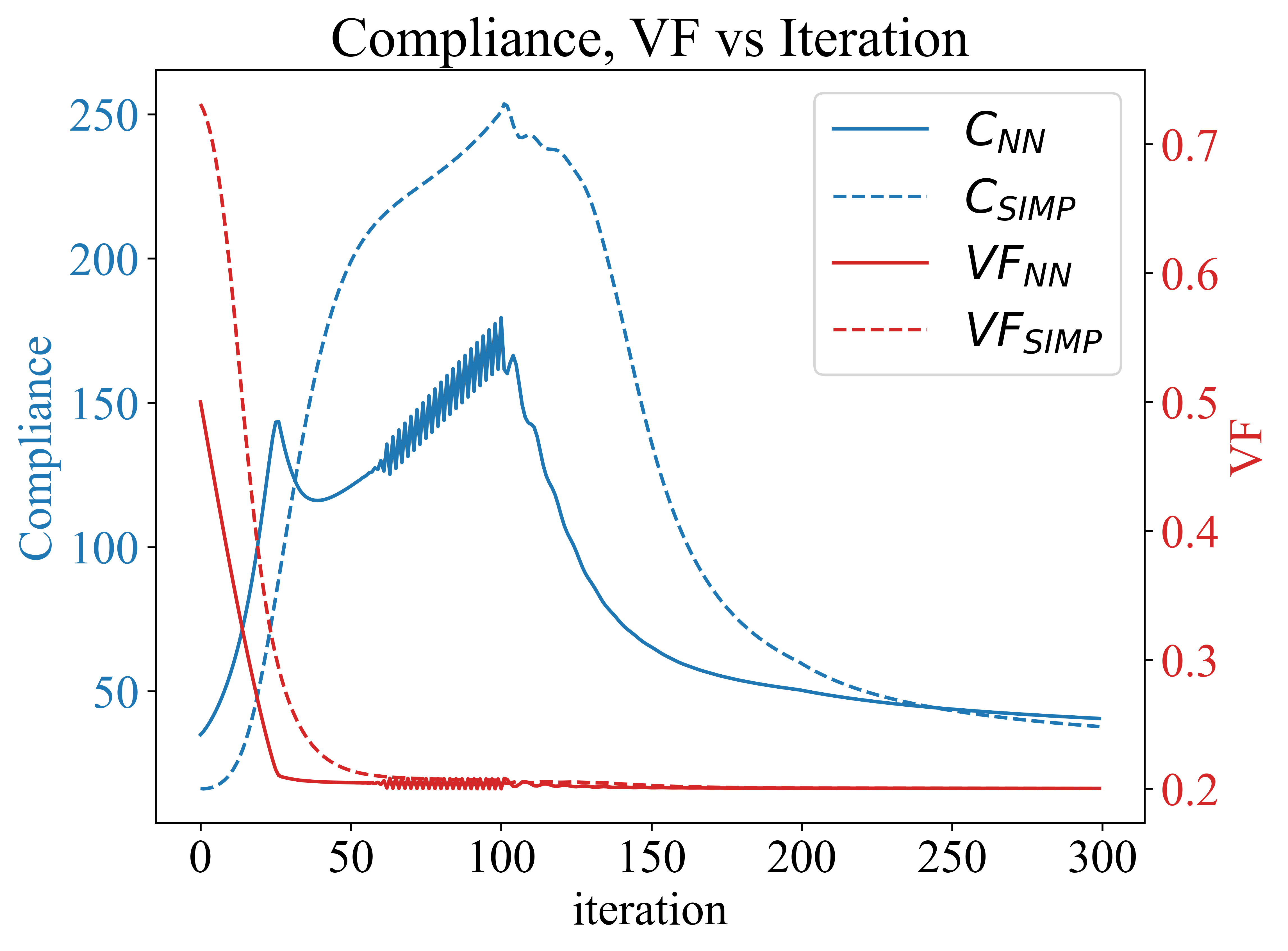}
\caption{}
\end{subfigure}

\centering
\caption{Comparing the topology and convergence history of neural network (b) and SIMP (c), the result is similar. (d) is the convergence history of neural network and SIMP. The oscillations between 60 to 100 epoch is due to using SGD as the optimizer for the first 100 epoch.}

\label{fig:simp_nn_compare}
\end{figure*}

\begin{figure*}
\centering

\begin{subfigure}[b]{0.3\textwidth}
\centering
\includegraphics[width=\textwidth]{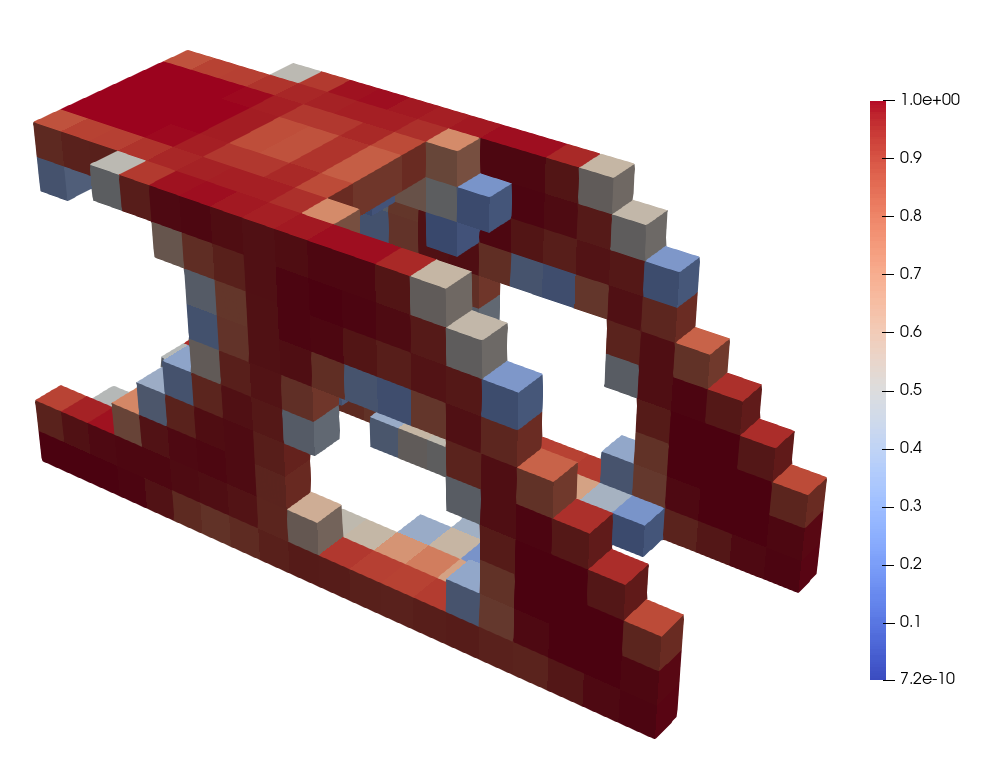}
\caption{Beam example at original resolution with neural network}
\end{subfigure}
\hfill
\begin{subfigure}[b]{0.3\textwidth}
\centering
\includegraphics[width=\textwidth]{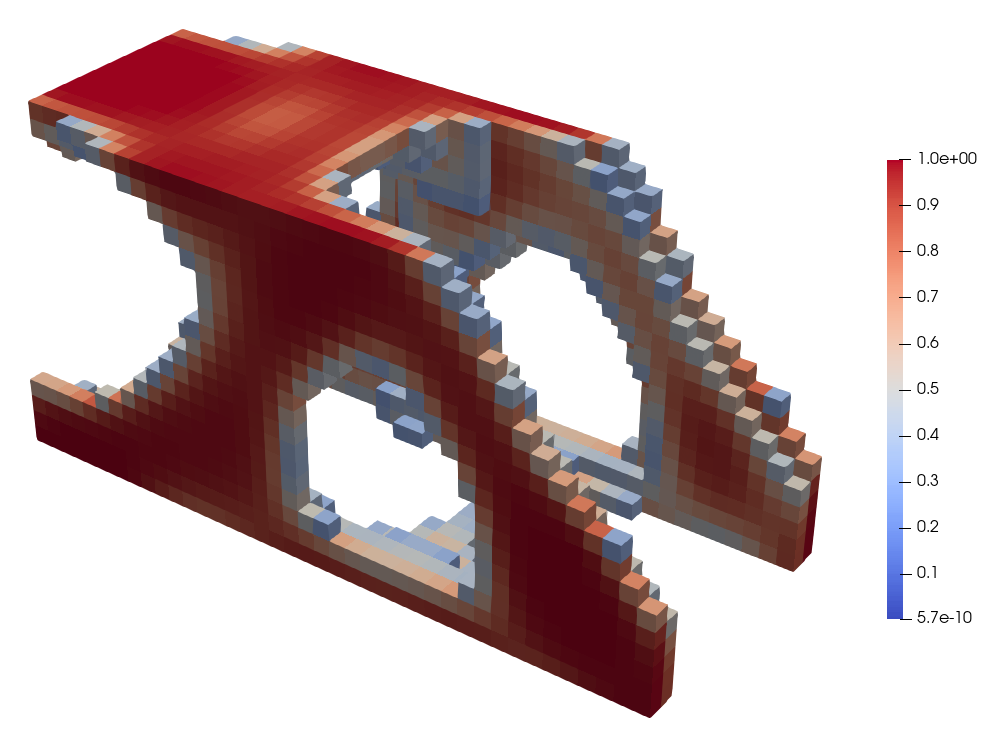}
\caption{Beam example 2$\times$ upsampled with neural network, $c=33.2$}
\end{subfigure}
\hfill
\begin{subfigure}[b]{0.3\textwidth}
\centering
\includegraphics[width=\textwidth]{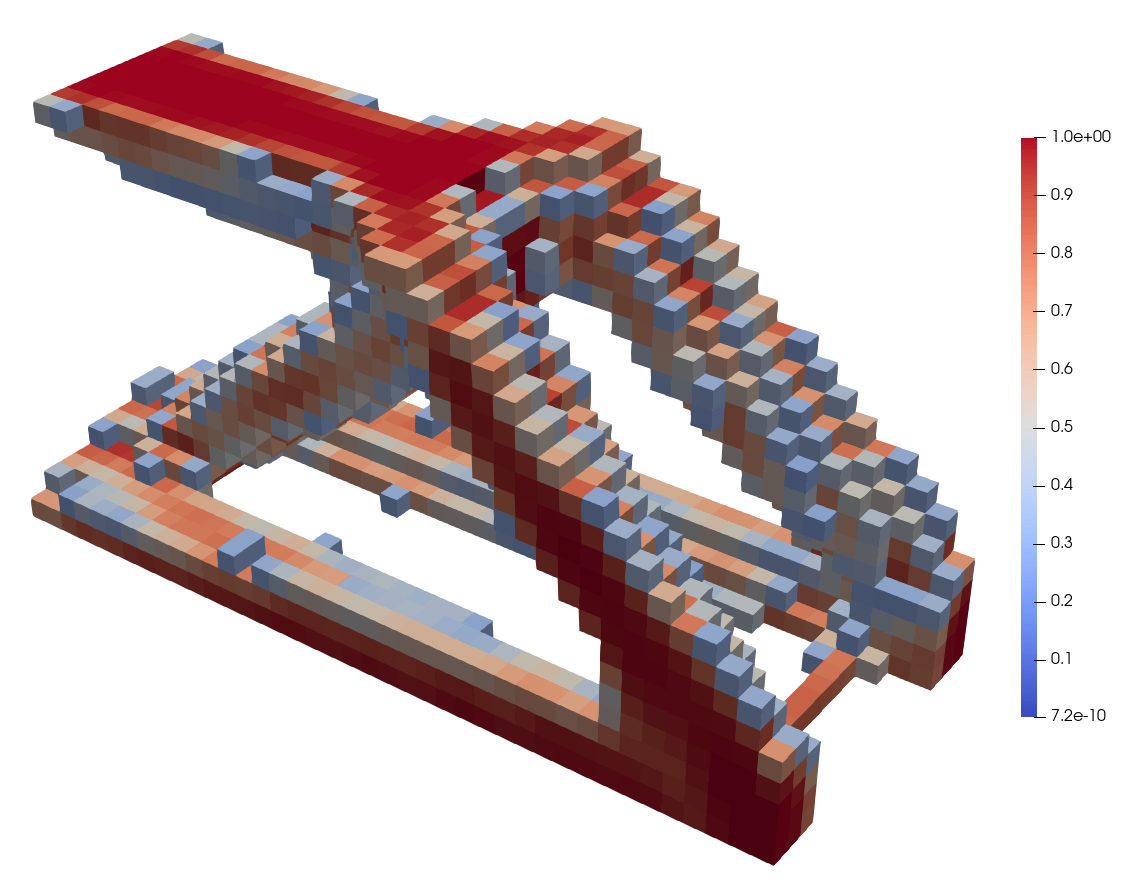} 
\caption{Beam example running at 2$\times$ the resolution, $c=20.9$}
\end{subfigure}

\centering
\caption{While running at higher resolution resulted in lower compliance, upsampling requires less computation time, enabling a tradeoff between optimality and speed.}

\label{fig:upsample_high_compare}
\end{figure*}
\begin{figure*}
\centering

\begin{subfigure}[b]{0.3\textwidth}
\centering
\includegraphics[width=\textwidth]{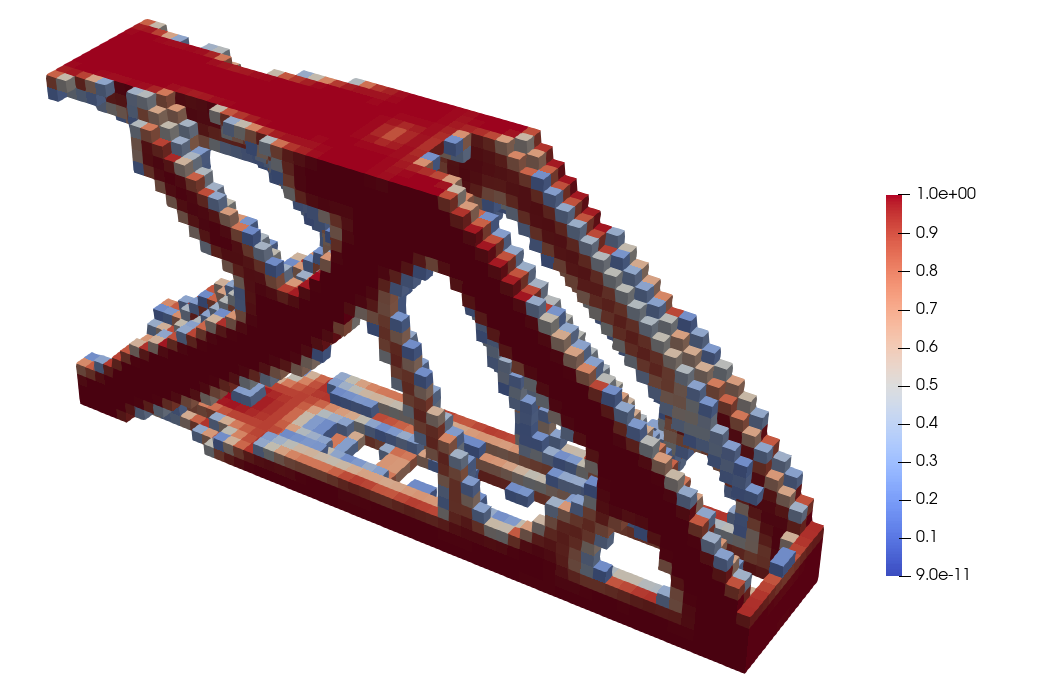}
\caption{c = 21.9}
\end{subfigure}
\hfill
\begin{subfigure}[b]{0.3\textwidth}
\centering
\includegraphics[width=\textwidth]{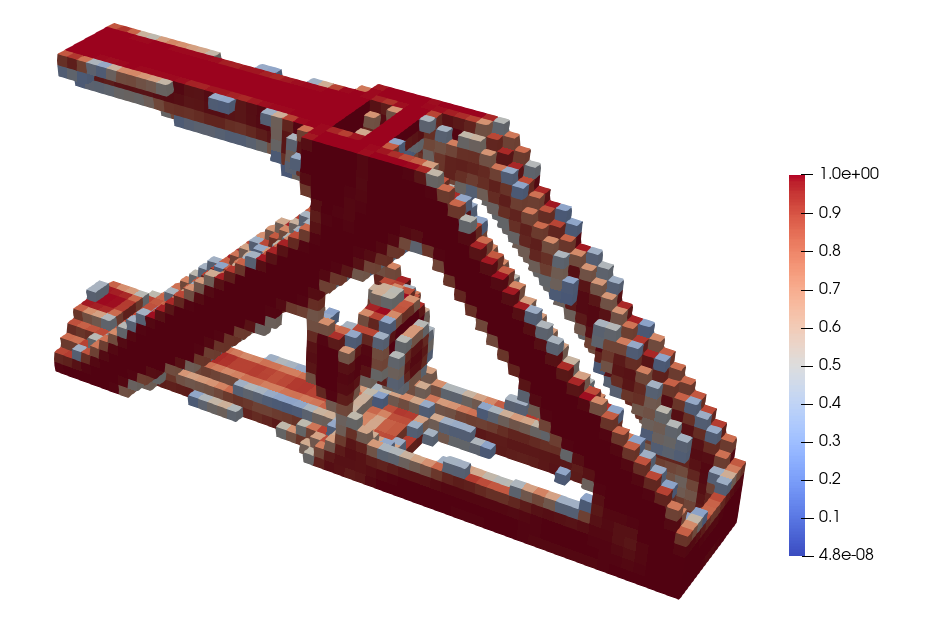}
\caption{c = 24.8}
\end{subfigure}
\hfill
\begin{subfigure}[b]{0.3\textwidth}
\centering
\includegraphics[width=\textwidth]{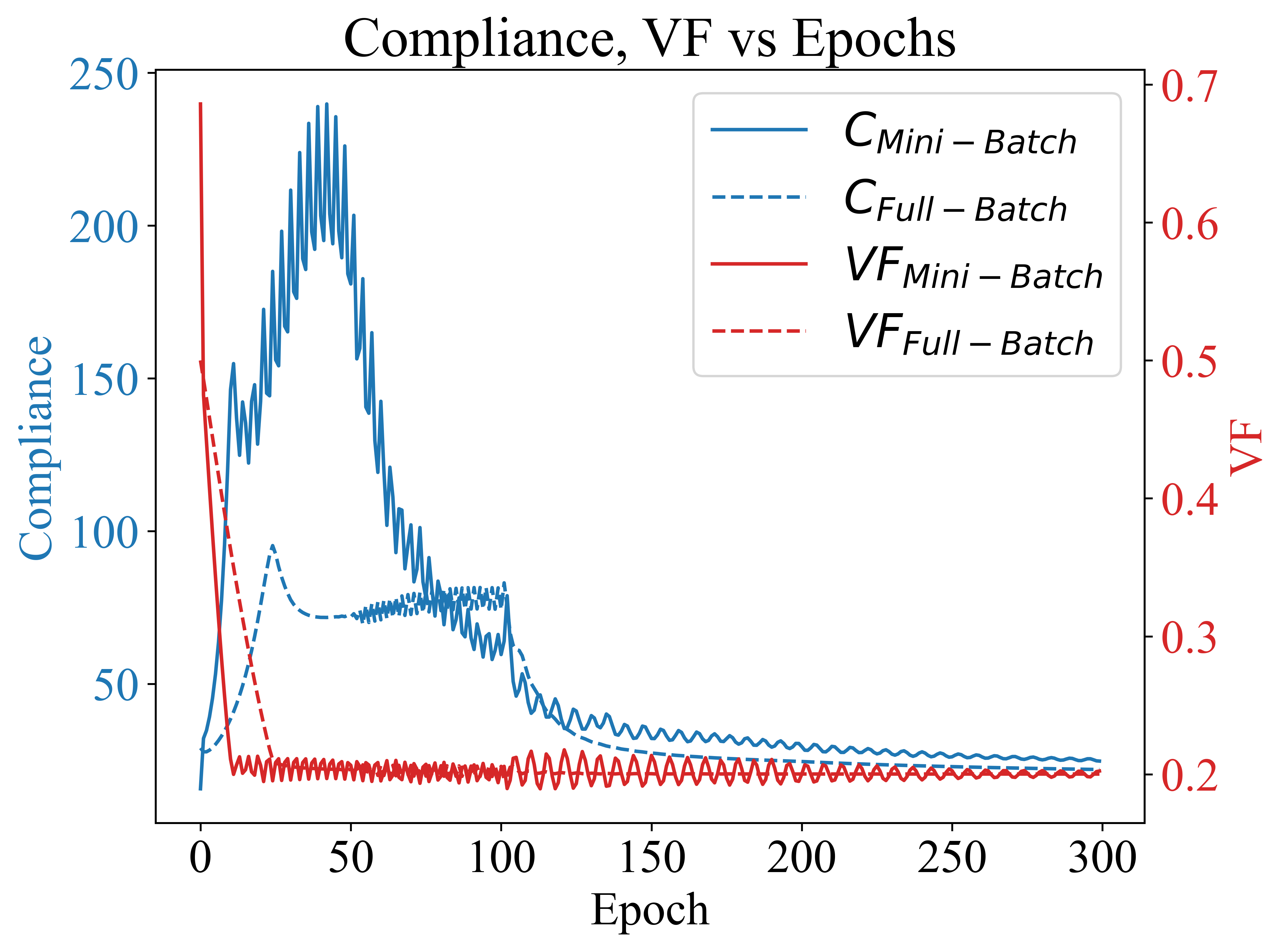}
\caption{}
\end{subfigure}

\centering
\caption{Comparing the result of (a) Full-batch optimization, (b) Mini-batch optimization, the topologies have small differences and compliance is slightly higher in mini-batch optimization. (c) is the convergence history comparison}

\label{fig:mini_batch}
\end{figure*}

\subsection{Topology Optimization}
During optimization, the topology network outputs the density value at the center for each element. These density values are then sent to the finite element solver to calculate compliance based on the SIMP interpolation. 
\begin{equation}
E_{i}(\rho_i) = E_{min}+\rho_i^{p}(E_{0}-E_{min})
\end{equation}
\begin{equation}
\textbf{K}\textbf{u} = \textbf{F}
\end{equation}
\begin{equation}
c = \textbf{F}^{T} \textbf{u}
\end{equation}

We only run 3D examples in this paper with 8 node hexahedral finite elements. We consider the material to be isotropic. For material anisotropy, a method similar to that described in \cite{orbayfukara2015jmd} can be utilized. The solver is based on the Matlab code from Liu and Tovar but we modified it to run on Python with CUDA accelerated sparse iterative matrix solver \cite{Liu2014}. The finite element solver is treated as a black box within the neural network. It takes in the density of each element and outputs the compliance and the sensitivity for each element with respect to the compliance. We also explore writing the finite element solver using only TensorFlow’s tensor representation. The automatic differentiation can correctly calculate the sensitivity but we choose to run the solver outside for computation speed purposes. 

Passive elements are non-designed regions in some problems. It has a density value of 0 to enforce it to be void. Usually, the 0 density values are directly imposed onto the finite element mesh \cite{Liu2014}. We employ a similar concept to the application of the continuous field by adding an additional passive loss to the total loss. The coordinates within the passive regions are first defined as $\textbf{X}_{passive}$. We then use equation \ref{eq_passive} to compute the L1 loss $\mathcal{L}_{passive}$ to force the density field within the passive region to be 0. Intuitively, $\textbf{X}_{passive}$ should always be smaller than $\textbf{X}$ which reduces the memory burden on the GPU.

\begin{equation}
\mathcal{L}_{passive}= \sum|T(\textbf{X}_{passive})-0|
\label{eq_passive}
\end{equation}

Symmetry can be enforced by mirroring the input coordinates across the symmetric axis. We then take the average density values from both the original $\textbf{X}$ and mirrored input $\textbf{X}_{mirrored}$. 
\begin{equation}
\rho_{sym} = 0.5(T(\textbf{X})+T(\textbf{X}_{mirrored}))
\end{equation}

\subsection{Optimization for additive manufacturing}
The first step for overhang detection is to calculate the density gradient $\nabla\rho$. We obtain the analytical solution of the density gradient from the input-output relation of the neural network. We then follow the proposed algorithm for overhang detection from Wang and Qian \cite{Wang2020}. We then add an additional height penalization to the overhang regions. 

We will briefly review the overhang detection proposed by Wang and Qian with a more detailed explanation of the additional height penalization. The overhang angle is defined as $\alpha$ which can be calculated from the boundary normal $\textbf{n}$ and the build direction $\textbf{b}$. The overhang angle $\alpha$ is defined as: 
\begin{equation}
\cos(\alpha) = \textbf{b}\cdot\textbf{n} = \textbf{b}\cdot\frac{\nabla\rho}{\|\nabla\rho\|_{2}}
\label{eq_overhang_def}
\end{equation}

The print angle network outputs angles in rotation around x and z-axis $R_x$ and $R_z$ respectively. An illustration for coordinate system and rotation definition is shown in Figure \ref{fig:sv_calc}(a). We use the following to convert it into build direction vector $\textbf{b}$
\begin{equation}
\textbf{b} = (\sin R_{z},\cos R_{x}\cos R_{z}, \sin R_{x}\cos R_{z})
\label{eq_overhang_def}
\end{equation}
For the printed structure to self-support, the critical overhang angle $\bar{\alpha}$ is defined (Figure \ref{fig:sv_calc}(b)). The lower bound of the overhang angle is constrained as $\alpha \geq \bar{\alpha}$. For every boundary region that is self supporting (Figure \ref{fig:sv_calc}(e)), it needs to satisfy
\begin{equation}
\textbf{b}\cdot\textbf{n} \leqslant \cos(\bar{\alpha})
\label{eq_overhang_ineq}
\end{equation}
Wang and Qian applied a Heaviside filter to $\textbf{b}\cdot\textbf{n} - \cos(\bar{\alpha})$ such that all boundary regions that do not satisfy the overhang constraint will be filtered positive. A smooth Heaviside function with $\beta$ set to 10 is used
\begin{equation}
H(\xi) = \frac{1}{1+e^{-2\beta\xi}}
\label{eq_heaviside}
\end{equation}
The overhang regions (Figure \ref{fig:sv_calc}(f)) can be integrated across the design domain to obtain the total overhang area
\begin{equation}
P_{\bar{\alpha}} = \frac{\int_{\Omega} H_{\bar{\alpha}} (\textbf{b},\nabla\rho) \textbf{b}\cdot\nabla\rho d \Omega} {\bar{A}}
\label{eq_overhang_sum}
\end{equation}
\begin{equation}
H_{\bar{\alpha}} (\textbf{b},\nabla\rho)= H(\textbf{b}\cdot \frac{\nabla\rho}{\|\nabla\rho\|_{2}} - \cos(\bar{\alpha}))
\label{eq_overhang_sum}
\end{equation}
$\bar{A}$ is the characteristic area and we set it to be the largest surface in the design domain. For sensitivity calculation, derivation is provided in Wang and Qian's paper \cite{Wang2020}. Since we implemented these calculations using TensorFlow's data structure, automatic differentiation will take care of sensitivity calculation. 

\subsection{Overhang formulation with height penalization}
We also add additional height penalization to Wang and Qian's method \cite{Wang2020}. Height penalization requires the detection of the lowest solid region within the design domain. In Wang and Qian's \cite{Wang2020} approach, overhang regions are split between the exterior boundary support and the interior boundary support to accommodate the sometimes necessary support on the lower side of the part. With our height penalization, necessary support on the underside is not penalized as much which reduces the need to split the overhang region calculation between the interior and exterior. The input coordinates are normalized between $-0.5$ and $0.5$ for the longest edge of the design domain. We first subtract $-1.0$ on the y-axis coordinates and then multiply with the Heaviside filtered density at each coordinate (Figure \ref{fig:sv_calc}(h)). 
\begin{equation}
\tilde{\rho} = \frac{1}{1+e^{-2\beta(\rho-0.5)}}
\label{eq_heaviside}
\end{equation}
\begin{equation}
\textbf{X}_{y,lowest} = min((\textbf{X}_y - 1.0)\tilde{\rho})+1.0
\label{eq_heaviside}
\end{equation}
During this step, any density value close to 0 will result in $(\textbf{X}_y - 1.0)\tilde{\rho}$ to be zero and filtered out. By taking the minimum of the multiplication, we can obtain the lowest region that has solid elements. We then multiply the $\textbf{X}_y$ coordinates with the overhang regions to obtain a height penalized overhang constraint (Figure \ref{fig:sv_calc}(i)). 
\begin{equation}
H_{\bar{\alpha}} = \frac{\int_{\Omega} H_{\bar{\alpha}} (\textbf{b},\nabla\rho) \textbf{b}\cdot\nabla\rho(\textbf{X}_y-\textbf{X}_{y,lowest})d \Omega} {\bar{A}}
\label{eq_overhang_sum}
\end{equation}


\begin{figure*}
\centering

\begin{subfigure}[t]{0.18\textwidth}
\centering
\includegraphics[width=\textwidth]{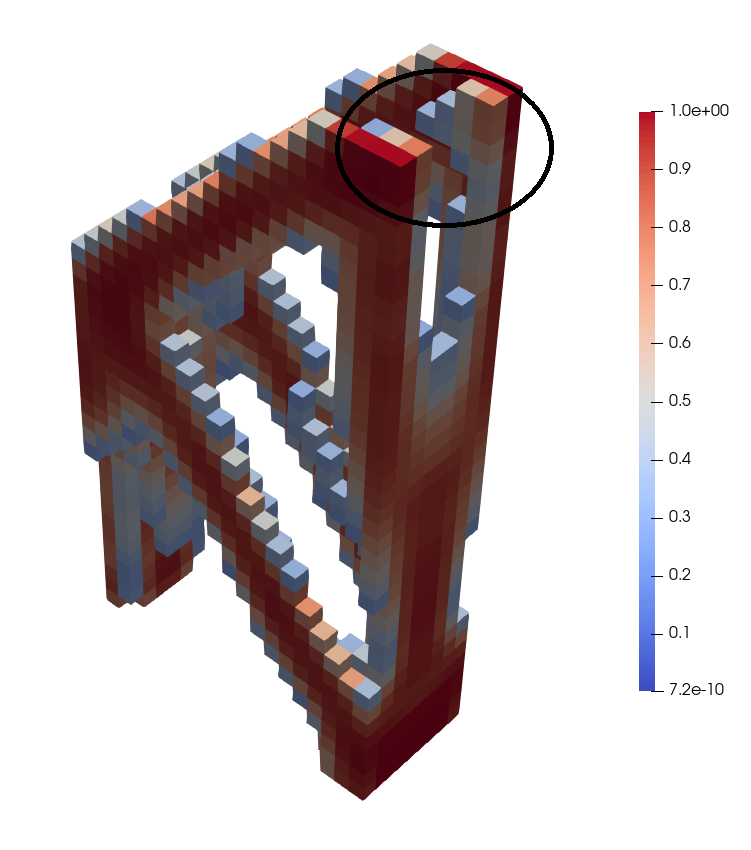}
\caption{$c=44.3$, $H_{\bar{\alpha}}=0.008$ }
\end{subfigure}
\hfill
\begin{subfigure}[t]{0.15\textwidth}
\centering
\includegraphics[width=\textwidth]{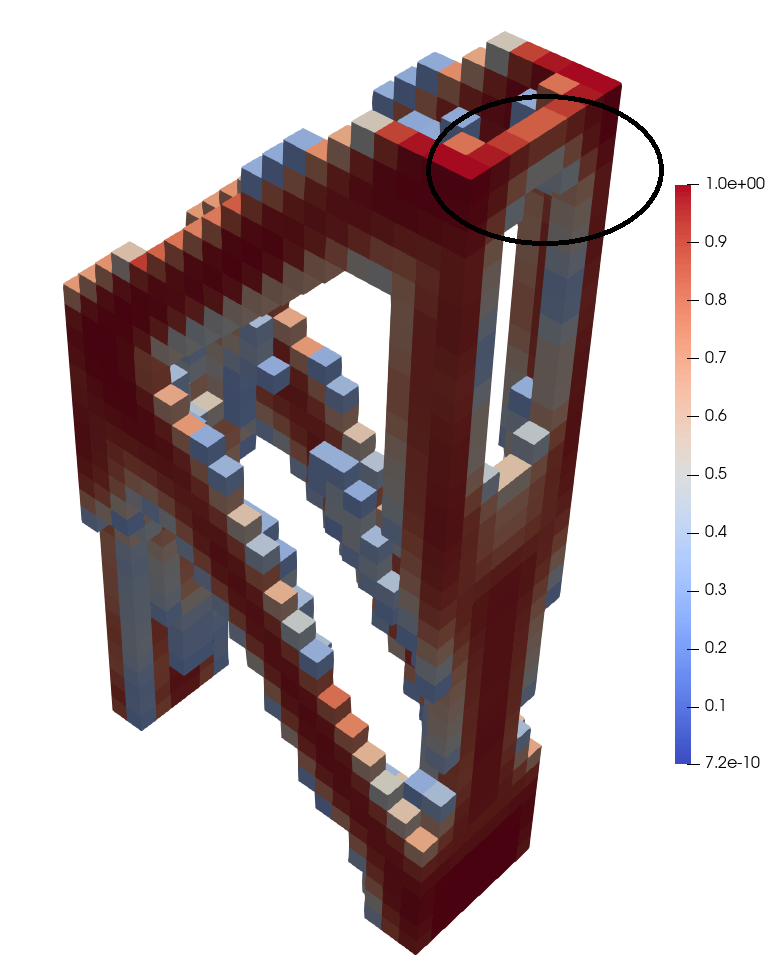}
\caption{$c=41.9$, $H_{\bar{\alpha}}=0.1278$}
\end{subfigure}
\hfill
\begin{subfigure}[t]{0.3\textwidth}
\centering
\includegraphics[width=\textwidth]{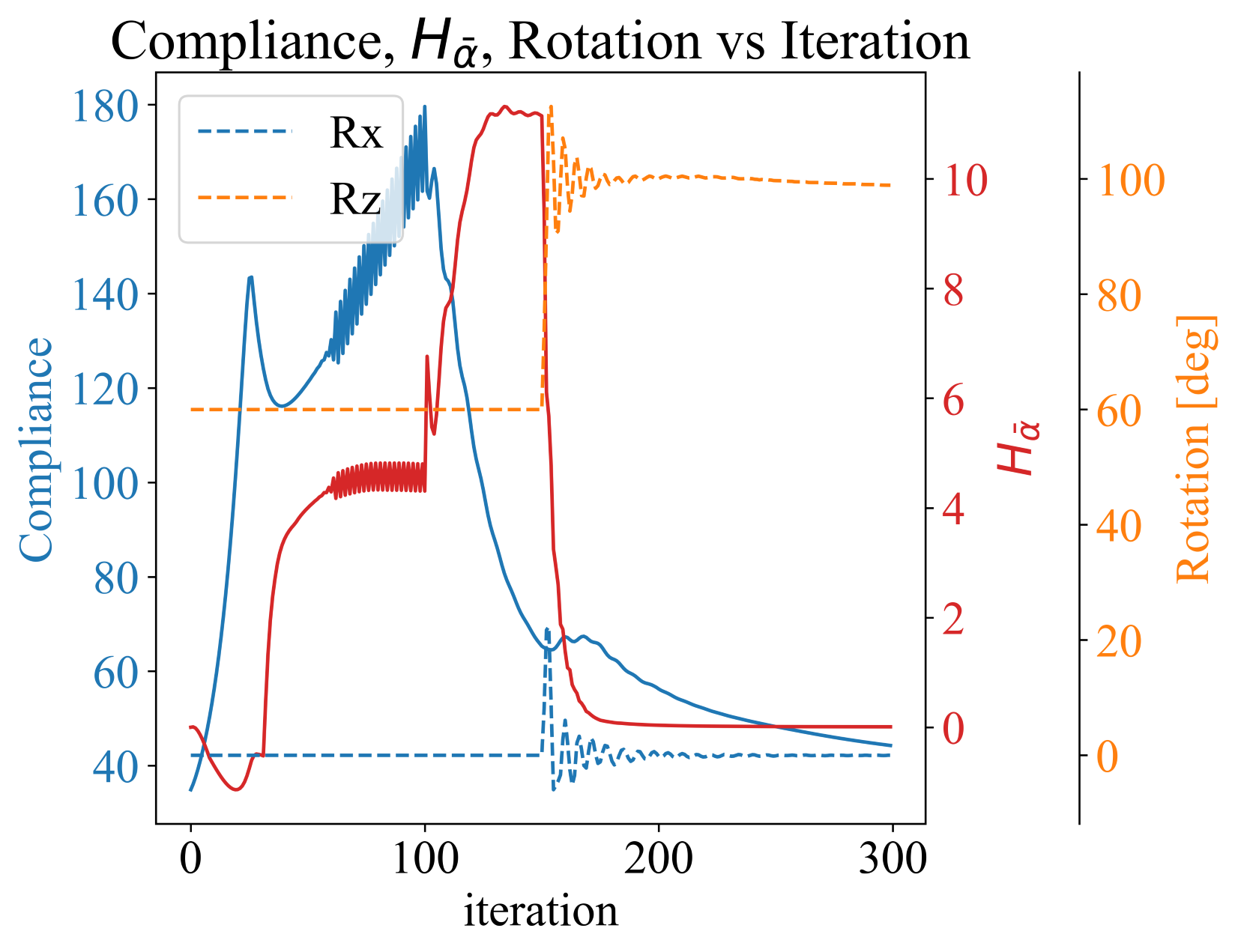}
\caption{}
\end{subfigure}
\hfill
\begin{subfigure}[t]{0.3\textwidth}
\centering
\includegraphics[width=\textwidth]{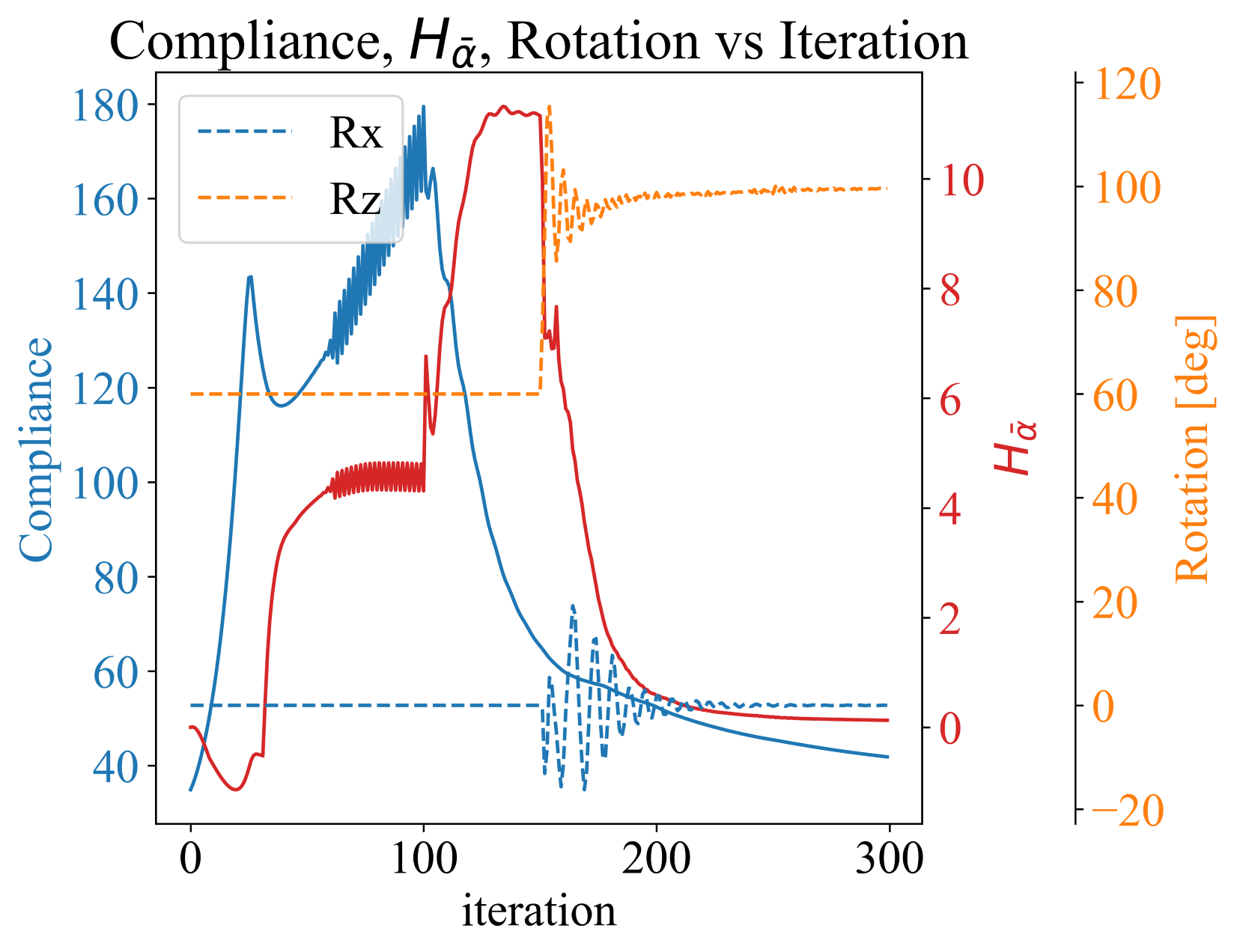}
\caption{}
\end{subfigure}

\centering
\caption{(a) and (b) are the voxel plot for with and without height penalization. With only overhang penalization, despite the geometry being optimized, still a noticeable overhang region is present on top (b). Adding the height penalization to overhang surfaces further reduced the overhang regions on top (a). (c) and (d) are the convergence history for with and without height penalization}

\label{fig:conc_beam}
\end{figure*}

\subsection{Loss function}
Both topology optimization and optimization for additive manufacturing is integrated into the machine learning framework. Variables that are being optimized are the weights of each neural network. Built-in optimizers such as Adam \cite{Abadi2016} and SGD are used to train the neural network. The constrained optimization problem needs to be transferred into unconstrained minimization problem for neural network. Chandrasekhar and Suresh \cite{Chandrasekhar2021} formulated the loss function for topology optimization including volume fraction and compliance constraint. We adopt the volume fraction and compliance constraint with the addition of overhang regions. The combined loss function is 
\begin{equation}
L = \frac{c}{c_0}+\alpha_1(\frac{\bar{\rho}}{V^*}-1)^2+\alpha_{2}H_{\bar{\alpha}}
\label{eq_overhang_sum}
\end{equation}
In the optimization, the volume fraction $V^*$ is an equality constraint. When $\alpha_1 \to \infty$, the equality constraint is satisfied. We assign a maximum value of 100 for $\alpha_1$ with initial value of 0 and gradually increase $\alpha_1$ every iteration. For the overhang constraint $H_{\bar{\alpha}}$, we do not target complete overhang removal. $\alpha_2$ gradually increases from 0 to 1 during training. For topology optimization, $c_0$ is the initial compliance calculated on the design domain with the uniform volume fraction $V^*$. 



\begin{figure*}
\centering

\begin{subfigure}[t]{0.25\textwidth}
\centering
\includegraphics[width=\textwidth]{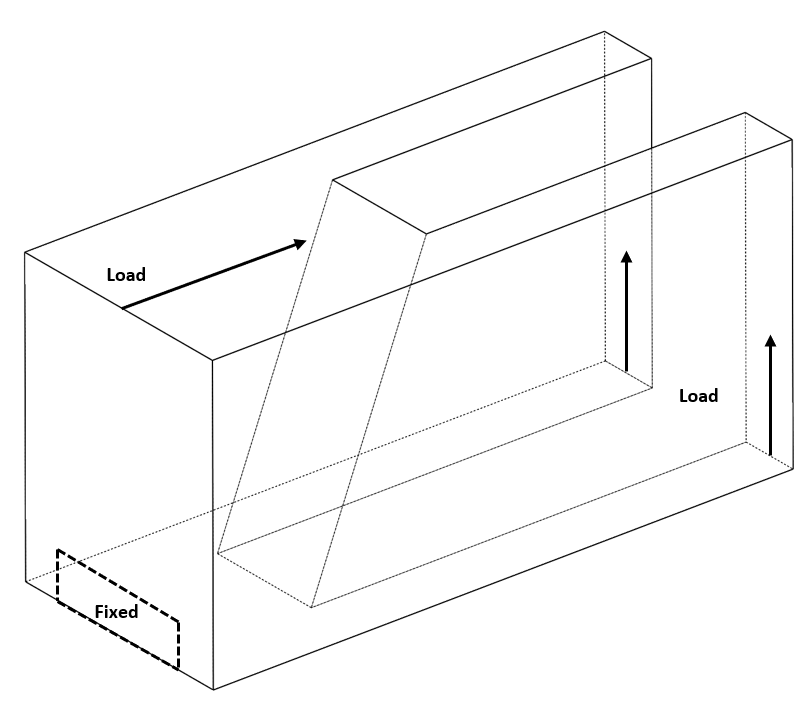}
\caption{Boundary condition for the bike frame}
\end{subfigure}
\hfill
\begin{subfigure}[t]{0.33\textwidth}
\centering
\includegraphics[width=\textwidth]{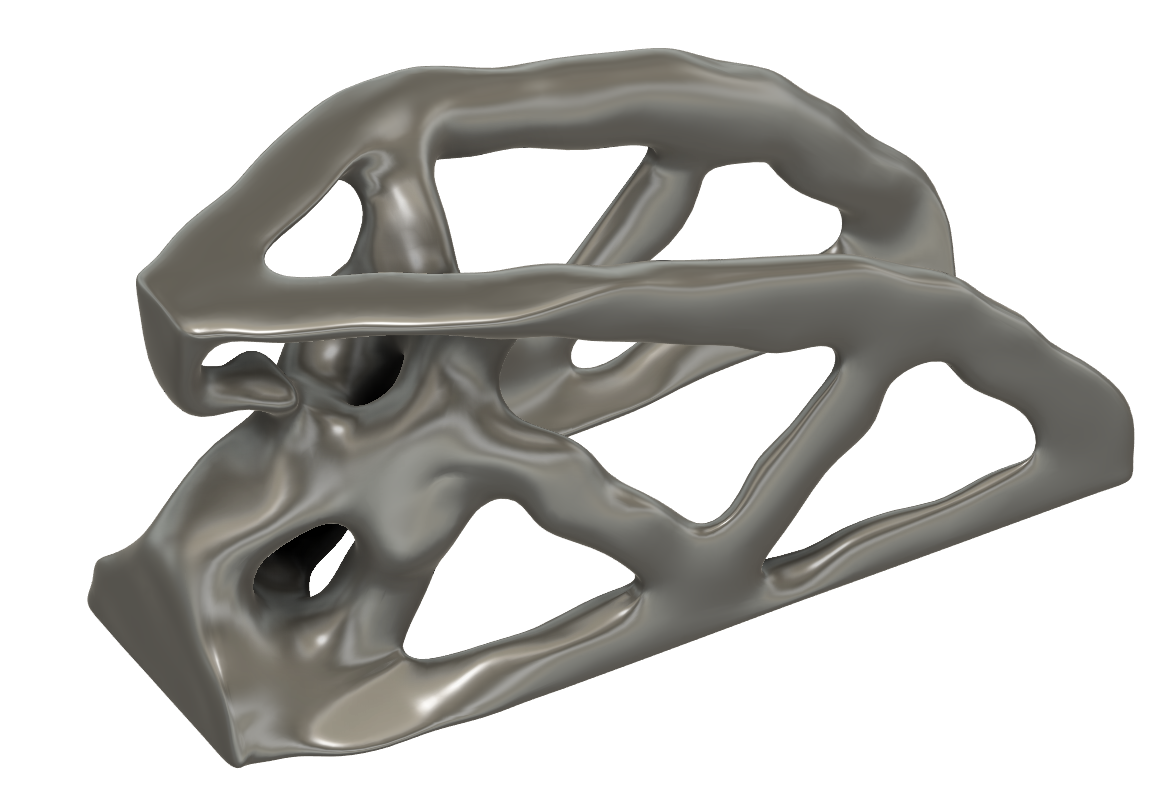}
\caption{Standalone topology optimization result, $c=116.4$, $H_{\bar{\alpha}}=8.229$}
\end{subfigure}
\hfill
\begin{subfigure}[t]{0.3\textwidth}
\centering
\includegraphics[width=\textwidth]{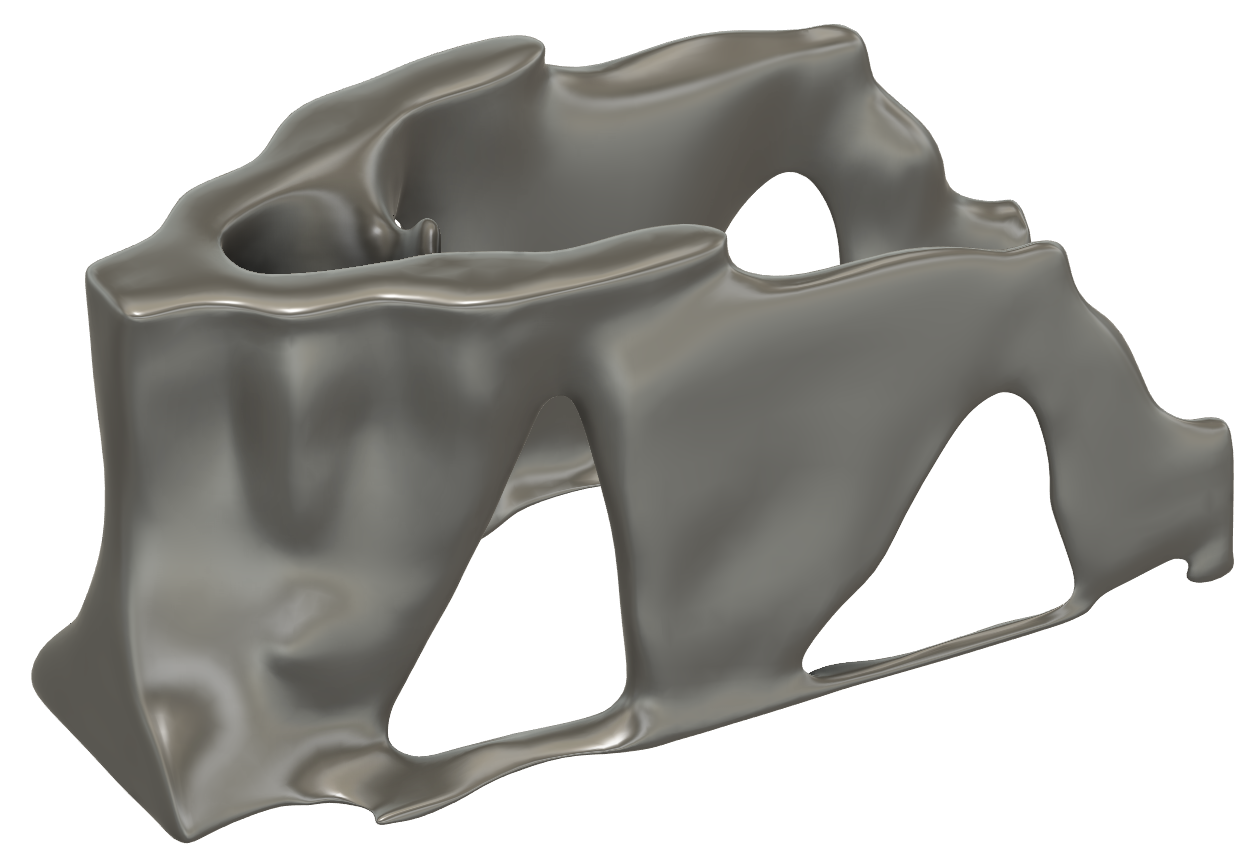}
\caption{ Optimized topology without segmentation $c=237$, $H_{\bar{\alpha}}=0.168$}
\end{subfigure}

\begin{subfigure}[t]{0.3\textwidth}
\centering
\includegraphics[width=\textwidth]{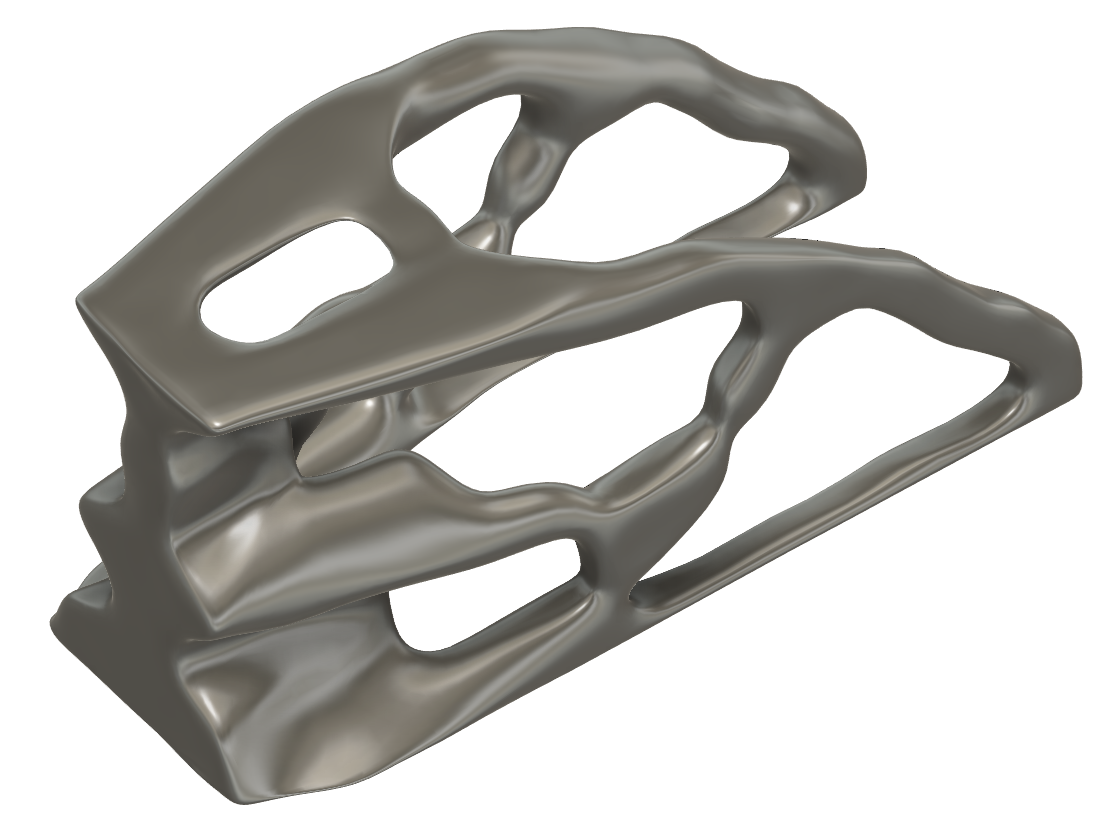}
\caption{Optimized topology with segmentation. $c=148$, $H_{\bar{\alpha}}=0.118$}
\end{subfigure}
\hfill
\begin{subfigure}[t]{0.3\textwidth}
\centering
\includegraphics[width=\textwidth]{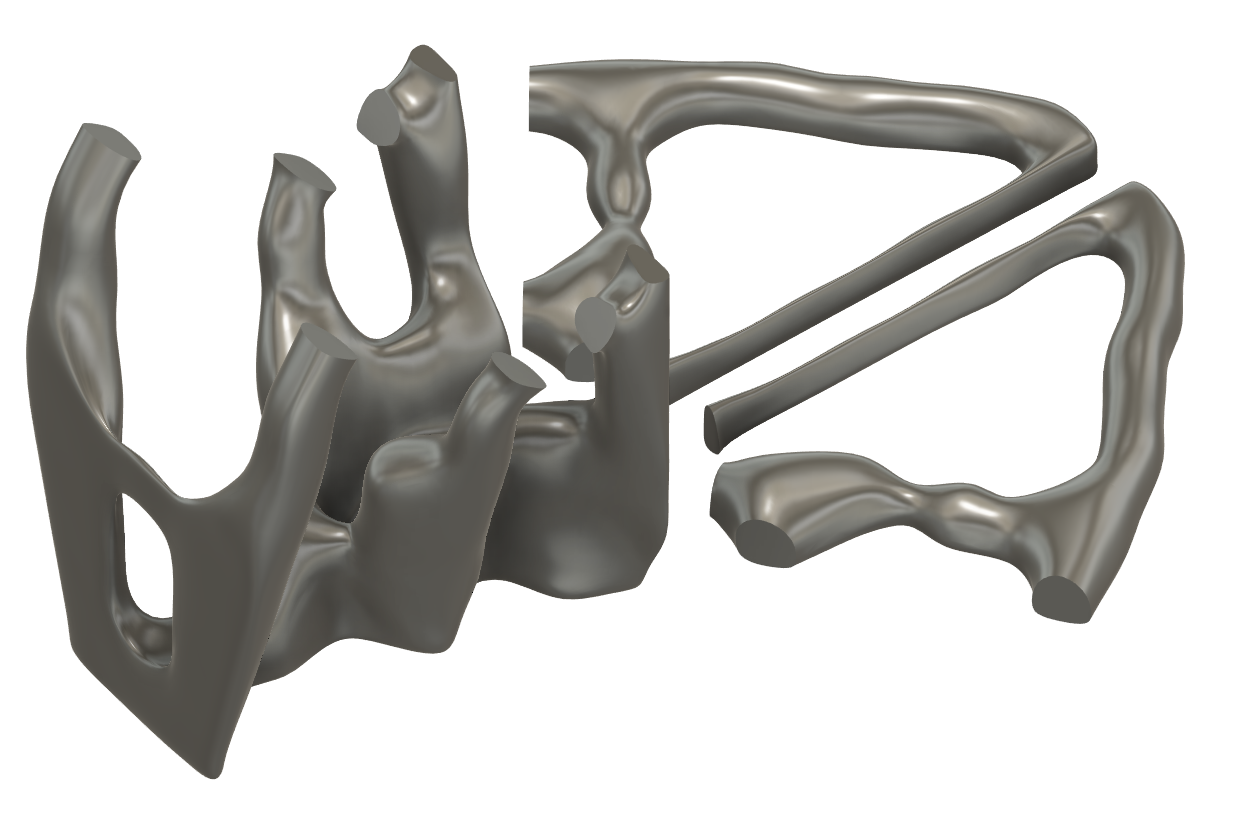}
\caption{Segmentation visualization for concurrent topology, print angle, and segmentation optimization}
\end{subfigure}
\hfill
\begin{subfigure}[t]{0.3\textwidth}
\centering
\includegraphics[width=\textwidth]{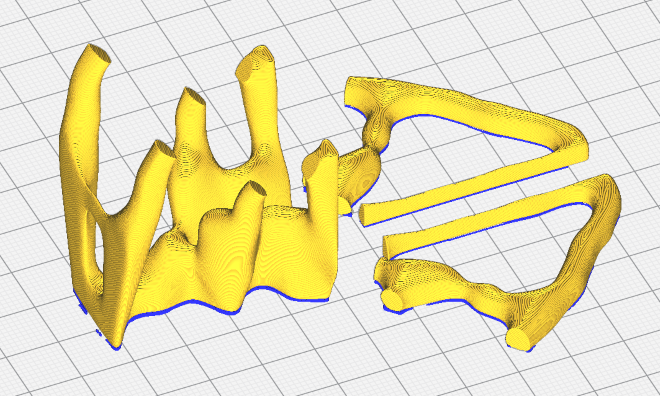}
\caption{Toolpath and support structure with segmentation}
\end{subfigure}

\begin{subfigure}[t]{0.6\textwidth}
\centering
\includegraphics[width=\textwidth]{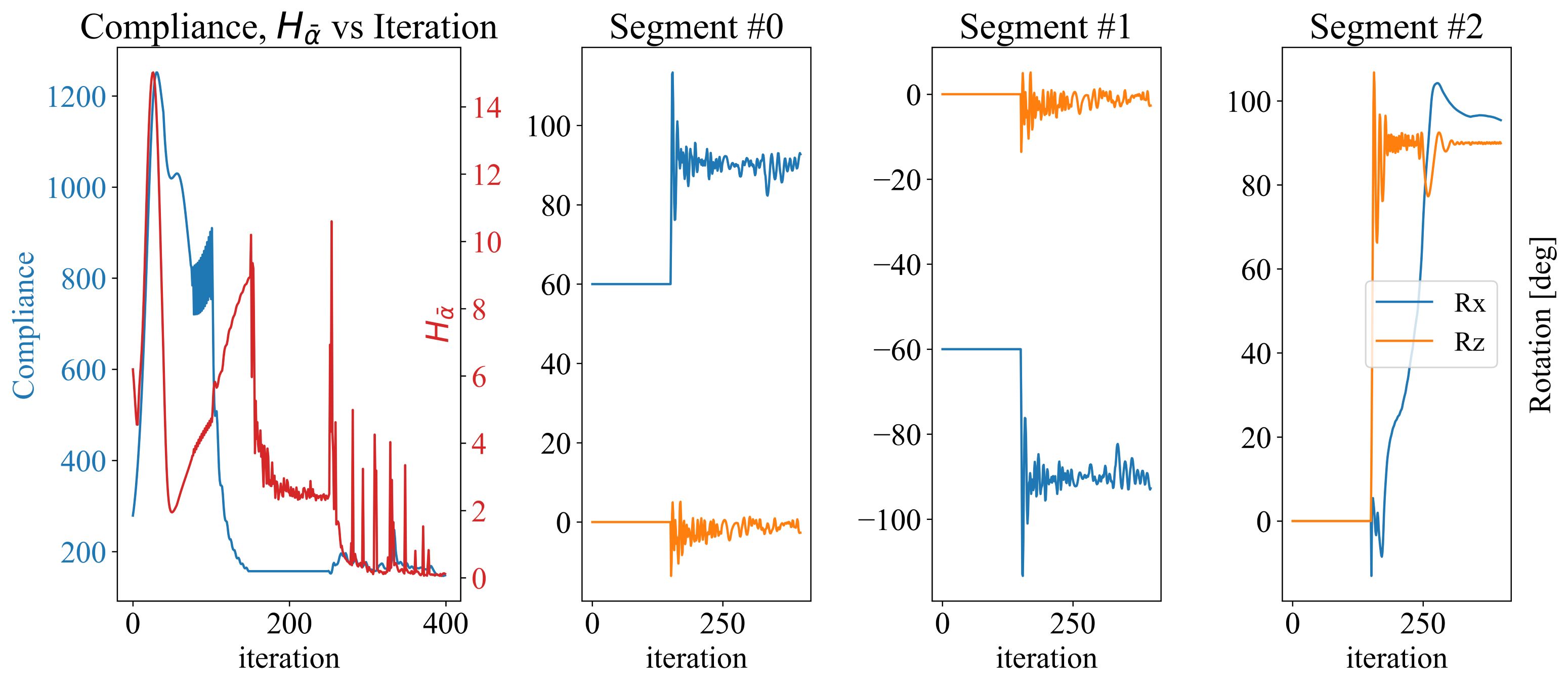}
\caption{Convergence history with segmentation enabled}
\end{subfigure}
\hfill
\begin{subfigure}[t]{0.35\textwidth}
\centering
\includegraphics[width=\textwidth]{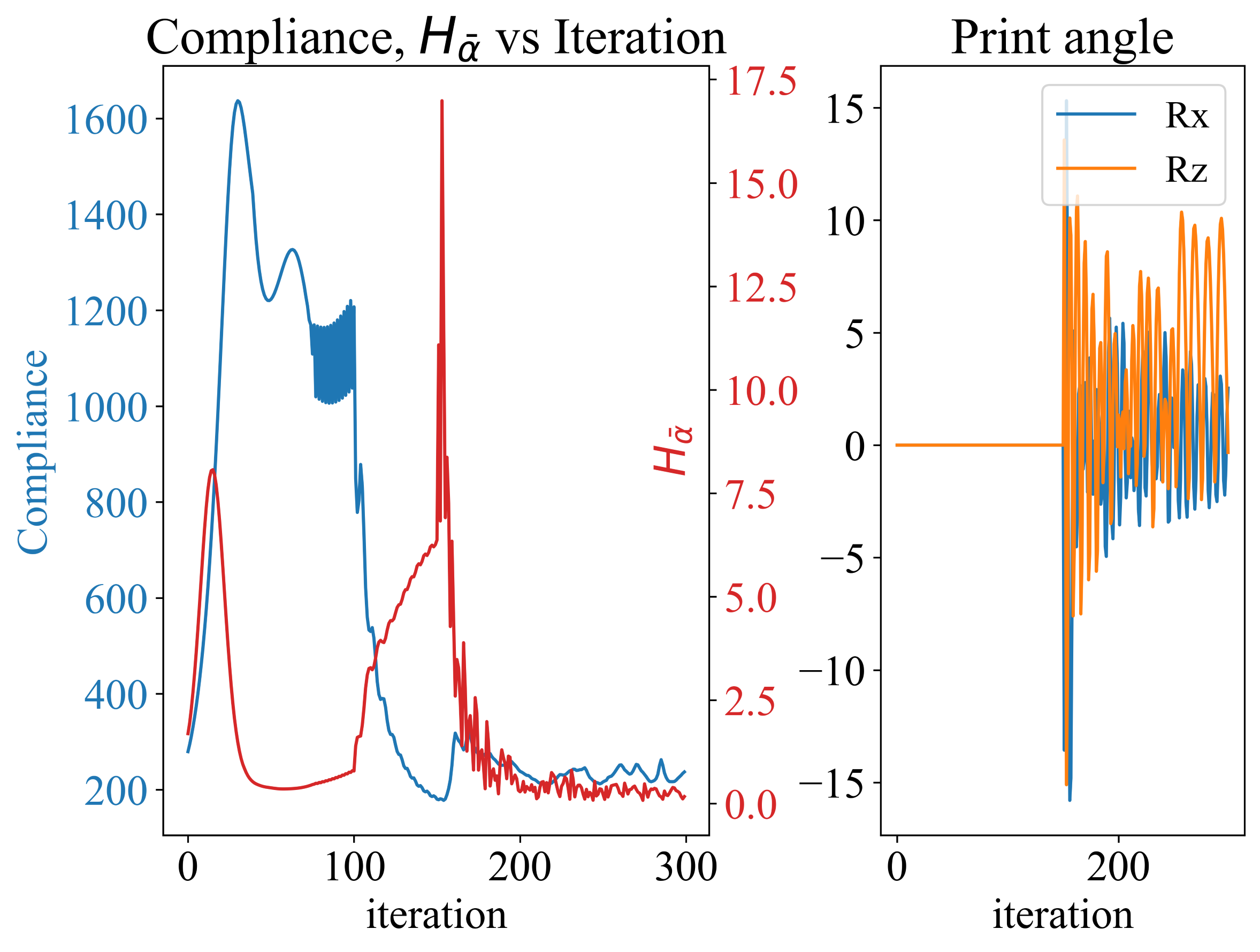}
\caption{Convergence history without segmentation}
\end{subfigure}
\centering
\caption{With segmentation, overhang regions can be further reduced while compliance is also lower since less change to topology is required. The gray geometry is obtained from using Top3dStl \cite{Liu2014} to convert voxelized result into stl file. }

\label{fig:frame_demo}
\end{figure*}

\section{RESULTS AND DISCUSSIONS}

\subsection{Validation}
We begin with comparing our neural network based optimization with SIMP. For the comparison, we setup SIMP using the same machine learning framework with the same cantilever beam boundary condition. The problem size is $40\times20\times8$ elements. We only enable the topology network within our modular architecture. The difference is that we configured a one layer neural network with 1 constant as input and the number of elements within the design domain as output. This configures the neural network to just store the optimization variables similar to the print angle network design. When running the optimization, for the first 100 iterations, we use SGD as the optimizer and afterward we switch to Adam \cite{Kingma2015}. We run the optimization for 300 iterations. 

We observe that compared to SIMP, the topology network converged to a similar compliance. We observe a  benefit of using a neural network to learn a continuous density field is that it inherently prevents checkerboard patterns without needing filtering (Figure \ref{fig:simp_nn_compare}(b)).

\subsection{Upsampling and mini-batch using the topology neural network}
With the continuous density field learned by the topology network, we can increase the resolution of the 3D coordinates $\textbf{X}_{upsample}$ by sampling at closer intervals within the original coordinates domain and obtain $\rho_{upsample}$ from the topology network. 
\begin{equation}
\rho_{upsample} = T(\textbf{X}_{upsample})
\end{equation}
We can then obtain a geometry at a higher resolution without the need for interpolation. This upsampled geometry is useful for exporting to be manufactured and/or running calculations at a higher resolution while still being differentiable. 

We compare the compliance and runtime between running at $20\times10\times8$ with $2\times$ upsampling with running at $40\times20\times16$ resolution (Figure \ref{fig:upsample_high_compare}). When comparing the runtime, running at the lower resolution takes 209 seconds while the higher resolution example takes 767 seconds, a 73$\%$ reduction in runtime. When upsampling the lower resolution to the same resolution, the compliance is calculated to be 59$\%$ higher. Running the optimization from a lower resolution and then upsampling it can be useful when a quick solution or preview of the optimization is desired while still maintaining acceptable structural performance.

There may be cases where topology optimization of high resolution geometries needs to be run. A mini-batch approach can be used here for alleviating the problem of high GPU memory usage. In the mini-batch approach, a small part of the geometry is updated in each epoch instead of the entire geometry, thus reducing GPU memory consumption. An example comparing the mini-batch and full-batch approaches is shown in Figure \ref{fig:mini_batch}, with the same boundary conditions as in the Figure \ref{fig:simp_nn_compare} example, but with a higher resolution geometry. The mini-batches are formed by splitting up the part evenly in the y-direction. 4 parts/batches are made in the given example, with each batch of 5400 voxels. The optimization is run for 300 epochs. The full-batch achieves a compliance of 21.94, while the mini-batch achieves a compliance of 24.77. Though the mini-batch optimization takes more time as FEA calculations for the entire geometry have to be performed for each batch, the maximum GPU memory required is significantly reduced, which for the given example is reduced from 5.74 GB to 2.74 GB.

\subsection{Overhang minimization}
With the promising result from running standalone topology optimization using only the topology network, we then enable the print direction network for simultaneous build direction and overhang optimization (Figure \ref{fig:conc_beam}(a)). With height penalization disabled, we can recreate the result from Wang and Qian albeit using a neural network to represent the topology with differentiable density gradients (Figure \ref{fig:conc_beam}(b)). We run two examples with both overhang region and print direction optimization enabled after 150 iterations to allow the geometry to first roughly converge. The critical overhang angle $\bar{\alpha}$ is set to be $45^{\circ}$. One has height penalization enabled while the other is disabled to compare the effect of height penalization. 

When comparing the result of these two examples, the compliance and optimal print direction are all similar. Both managed to significantly reduce the total overhang area. A noticeable distinction is that with the additional height penalty, the cantilever on the top has been removed (Figure \ref{fig:conc_beam}(a)). Despite being a very small overhang, its location is further away from the build plate which was removed with the addition of height penalization. When comparing the height-adjusted overhang penalization, our method manage a 94$\%$ decrease in the total overhang regions. 

\subsection{Concurrent build direction, part segmentation and topology optimization}
In the previous example, we demonstrated the effect of adding build direction into the optimization routine. We would like to further reduce the support structure for complex designs. We choose a simplified bike frame for this example. The boundary condition is shown in Figure \ref{fig:frame_demo}(a). Symmetry constraint is enforced on the design. We run three experiments with (1) standalone topology optimization, (2) simultaneous print angle and topology optimization and (3) 3 segments with concurrent print angle and topology optimization. For the first 150 iterations, only topology optimization is enabled to obtain a rough geometry. Then for the experiment with segmentation enabled, we run 100 iterations with only print angle optimization enabled. This step is to prevent the topology and segmentation from converging faster than the print angle resulting in optimization being stuck at a suboptimal print angle. At 250 iterations, build direction, part segmentation, and topology optimization is enabled.

When comparing the result of the three experiments, we observed that with segmentation(Figure \ref{fig:frame_demo}(d)), the overhang is reduced by 30$\%$ while compliance is smaller compared to without segmentation (Figure \ref{fig:frame_demo}(c)). Comparing to standalone topology optimization, our concurrent optimization managed to reduce overhang by 98.5$\%$. With segmentation, the two segments in the rear can be laid flat on the build plate with fewer restrictions on its topology to further reduce overhang (Figure \ref{fig:frame_demo}(d)). While there is some small amount of overhang on the bottom of the front segment, the optimization still rotated the part such that support higher up the build plate is eliminated. Without segmentation, the second experiment also has overhang regions reduced but has higher compliance and more overhang (Figure \ref{fig:frame_demo}(c)). Due to the geometry of the bike frame, print angle optimization alone has limited effect.

To explore the relationship between overhang and support structure, we exported the geometry as stereolithography files and imported them into Cura with the optimized print direction applied \cite{Ultimaker2020}. We set the overhang angle to be $45^\circ$. The part is shown in yellow with support structures shown in blue. We observe that almost all support material has been removed.

\section{LIMITATIONS AND FUTURE WORK}
One concern with segmented parts is that it needs welding or assembly after printing. However, large components require segmentation to fit inside the build chamber already. Cost may also be a consideration to decide if the support structure reduced is worth the additional effort of piecing parts together. Nevertheless, we demonstrated with the addition of segmentation, support structures can be further reduced. 

The present work explored optimization for additive manufacturing. In the future, we plan to further develop upon the current framework by adding cost models to conduct optimization for manufacturing costs. With the cost model added, we can explore the cost related to the number of segments by accommodating the assembly cost. Currently, we treat the mating surface between each segment with the same mechanical performance while welding and assembly may alter the mechanical performance. Furthermore, subtractive machining constraints may be added from leveraging the continuous density field by the neural networks.

\section{CONCLUSIONS}
We present a framework for concurrent build direction, part segmentation, and topology optimization using neural networks that aims to reduce support structures. With only the topology network, we benchmark against SIMP and demonstrate that similar results can be obtained from using a neural network to learn the density field of topology optimization. When the print direction network is enabled, simultaneous print direction and topology optimization can be achieved. Detecting overhang regions and minimization can be done effectively using the analytical density gradient obtained from the input-output relationship of neural networks. Finally, with a bike frame example, we demonstrated the effectiveness of combined build direction, part segmentation, and topology optimization in reducing support structures. 

\section*{Acknowledgment}
This research was funded by Army  Research Laboratory 50114.5.9.1130260. We would like to thank Krishnan Suresh and Aaditya Chandrasekhar for their insightful feedback. We also thank Sangjin Jung for help in problem formulation, data generation, and validation.

\bibliographystyle{unsrtnat}
\bibliography{References}








\begin{figure*}
\centering

\begin{subfigure}[t]{0.2\textwidth}
\centering

\includegraphics[width=\textwidth]{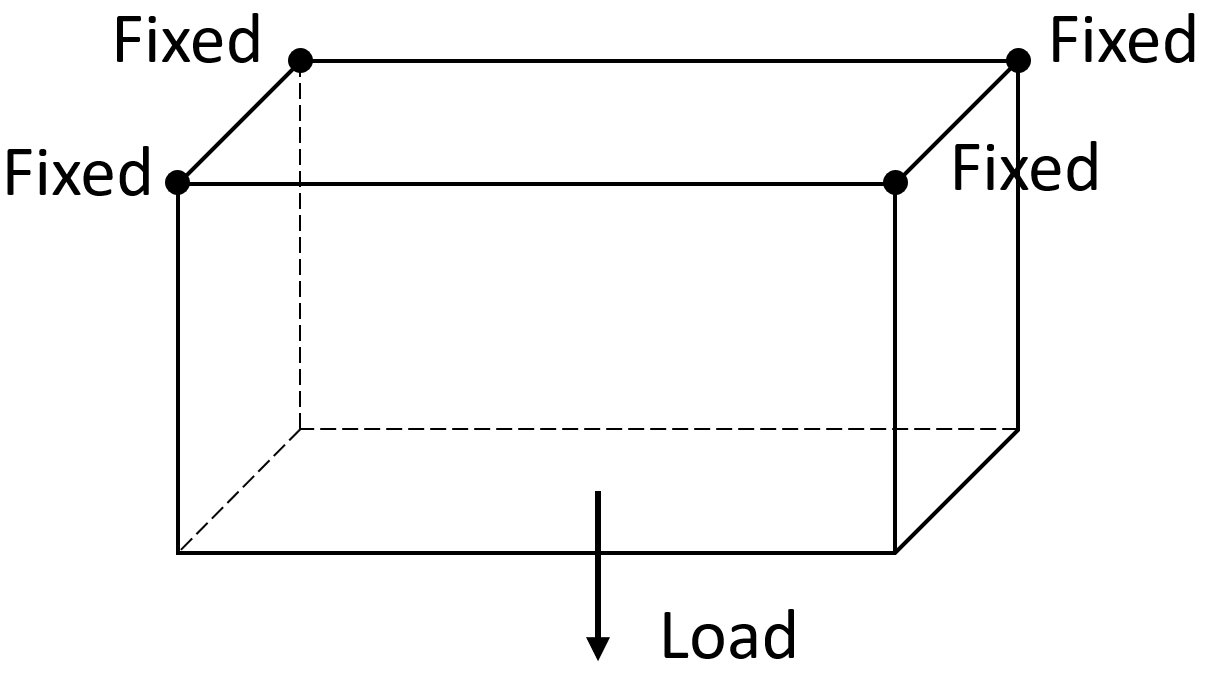}
\vspace{1.0cm}
\vfill
\centering
\includegraphics[width=\textwidth]{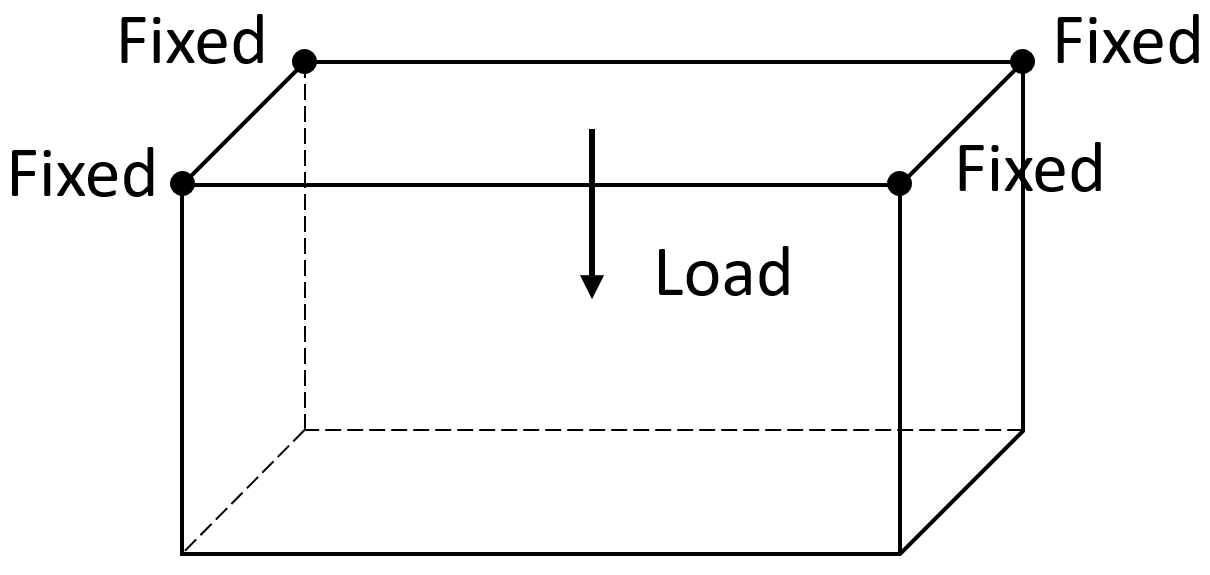}
\vspace{1.0cm}
\vfill
\centering
\includegraphics[width=0.8\textwidth]{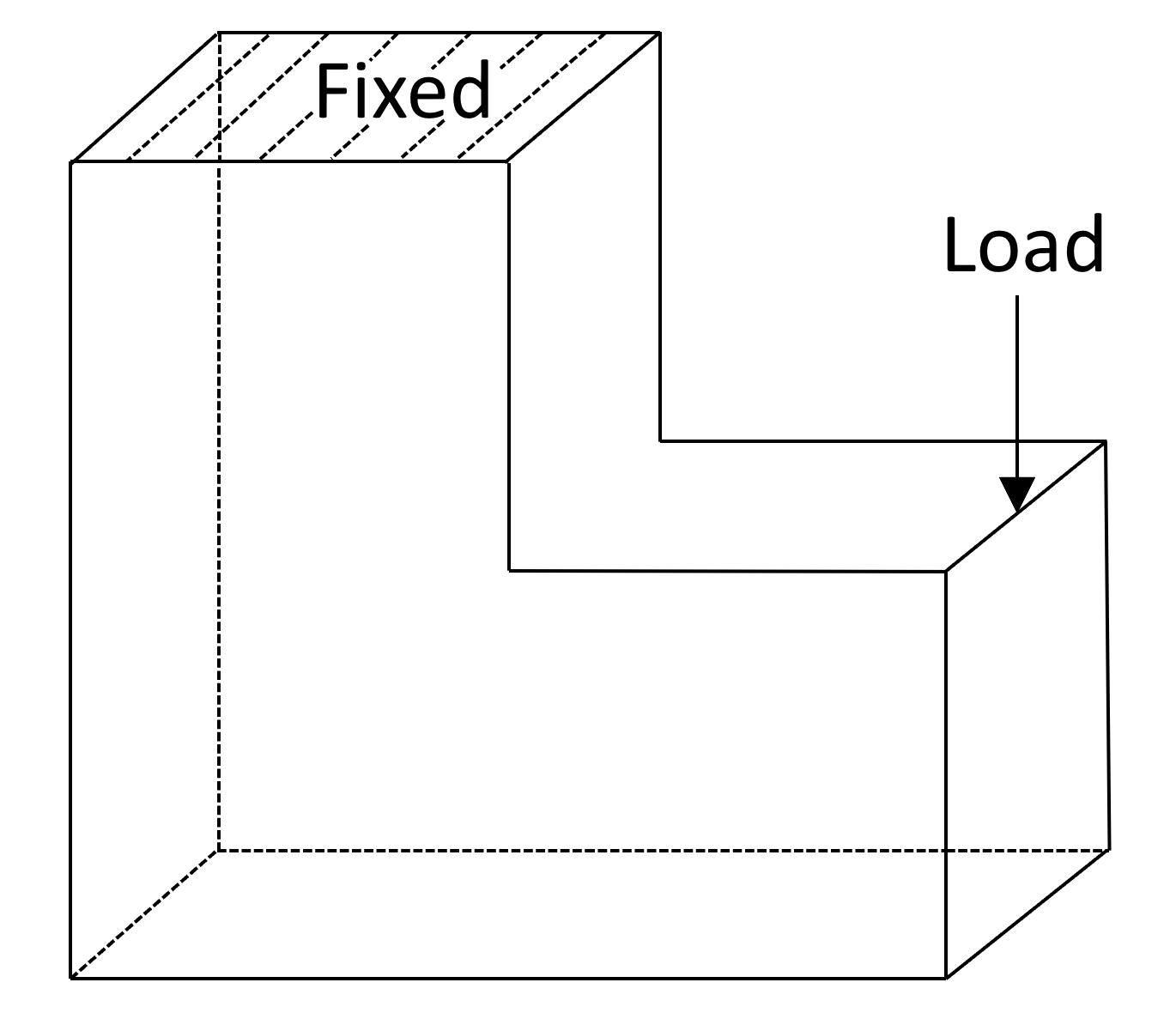}
\vspace{1.0cm}
\caption{}
\end{subfigure}
\qquad
\begin{subfigure}[t]{0.15\textwidth}
\centering
\includegraphics[width=\textwidth]{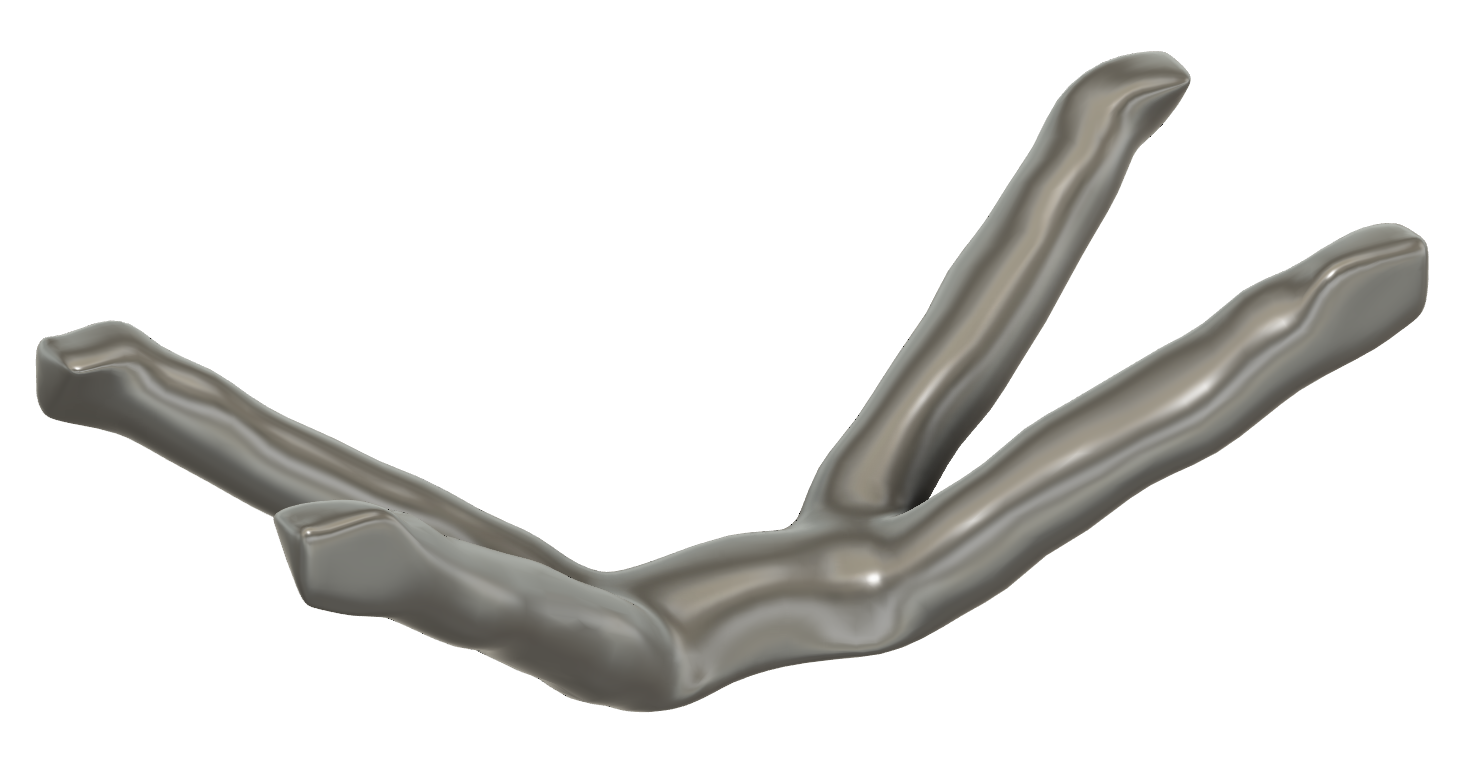}
$c=14$,
$H_{\bar{\alpha}}=3.843$
\vspace{0.5cm}
\vfill
\centering
\includegraphics[width=\textwidth]{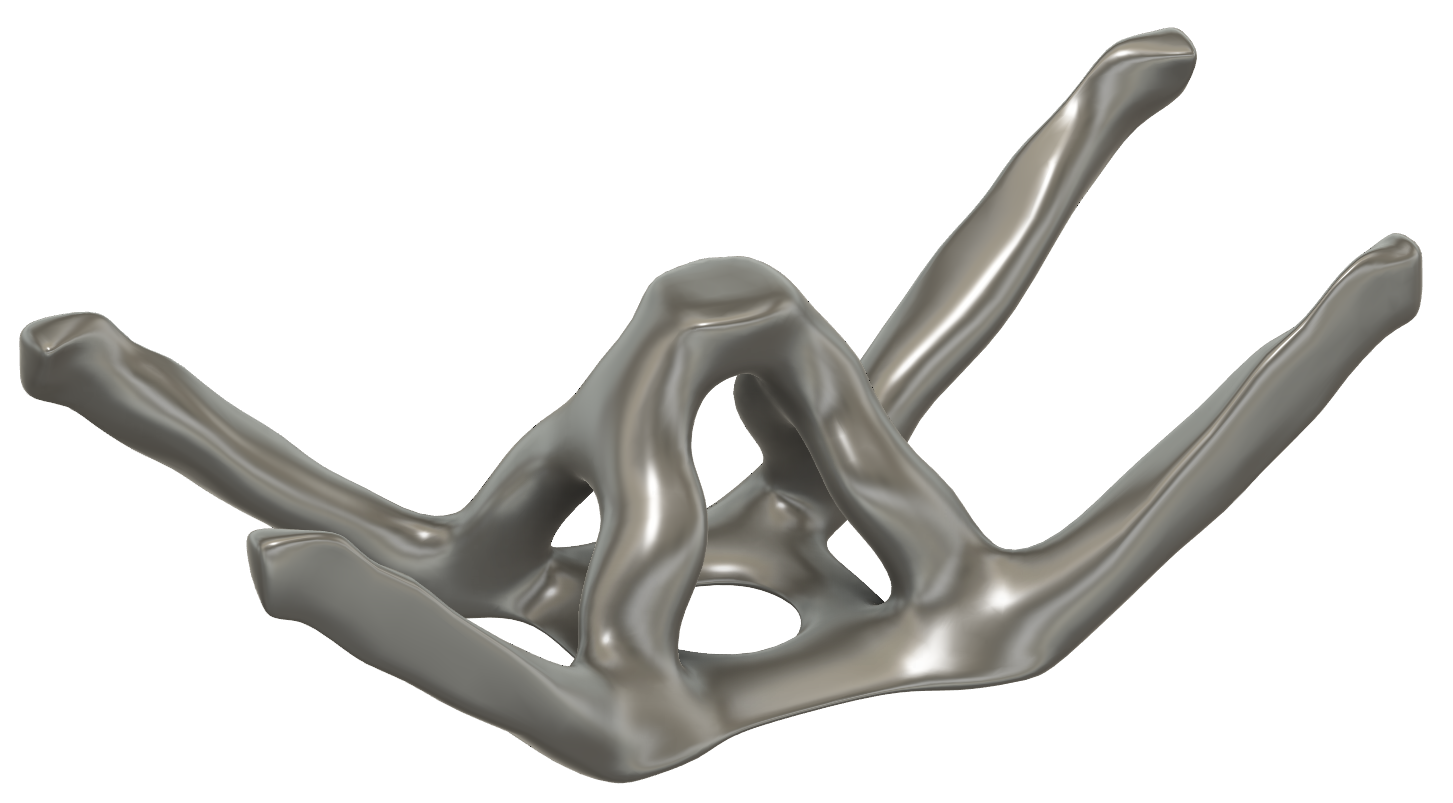}
$c=23$,
$H_{\bar{\alpha}}=2.926$
\vspace{0.5cm}
\vfill
\centering
\includegraphics[width=\textwidth]{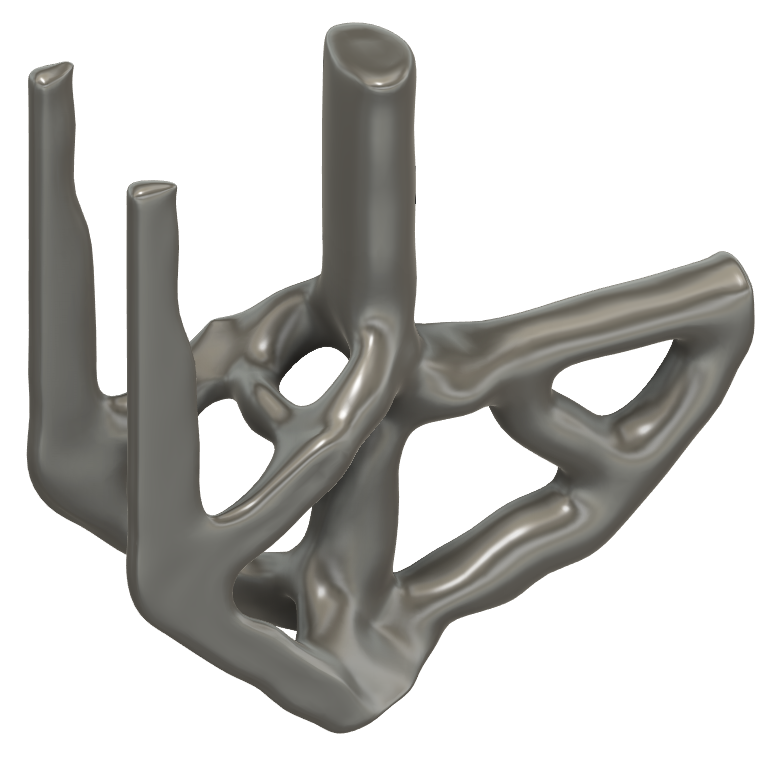}
$c=45$,
$H_{\bar{\alpha}}=2.668$
\vspace{0.2cm}
\caption{}
\end{subfigure}
\qquad
\begin{subfigure}[t]{0.15\textwidth}
\centering
\includegraphics[width=\textwidth]{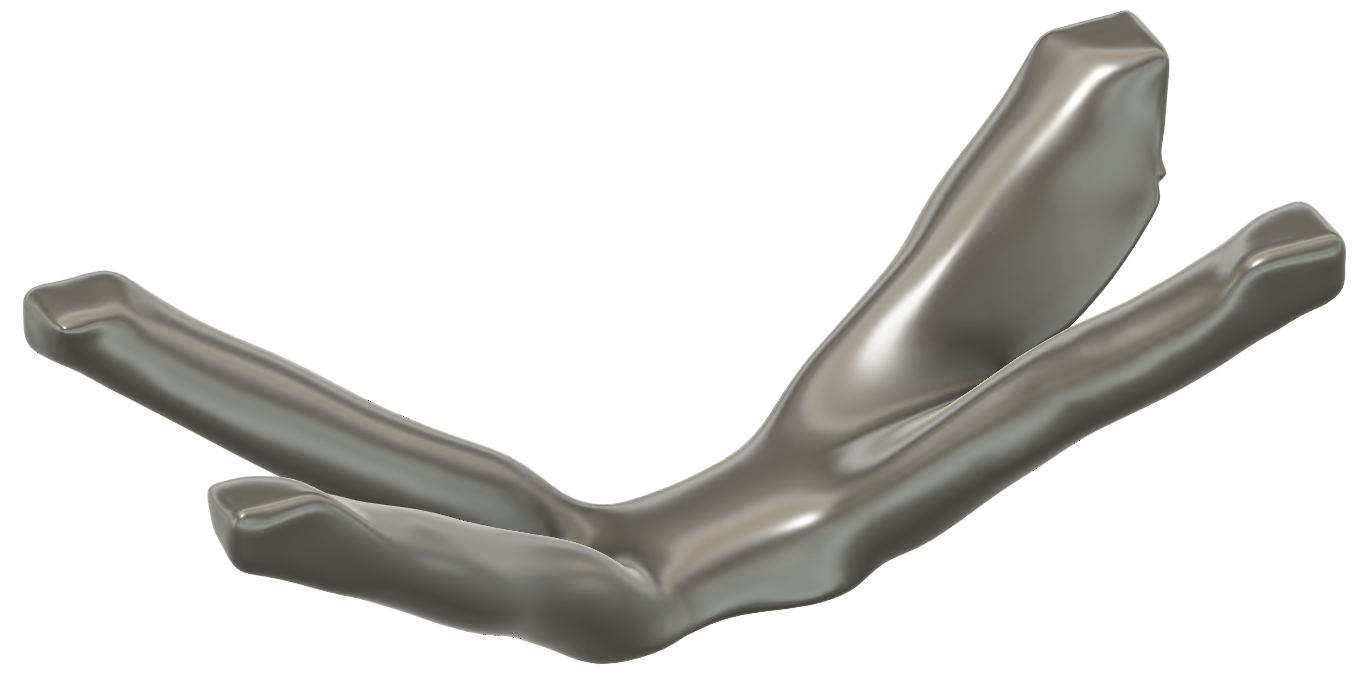}
$c=15$,
$H_{\bar{\alpha}}=0.074$
\vspace{0.6cm}
\vfill
\centering
\includegraphics[width=\textwidth]{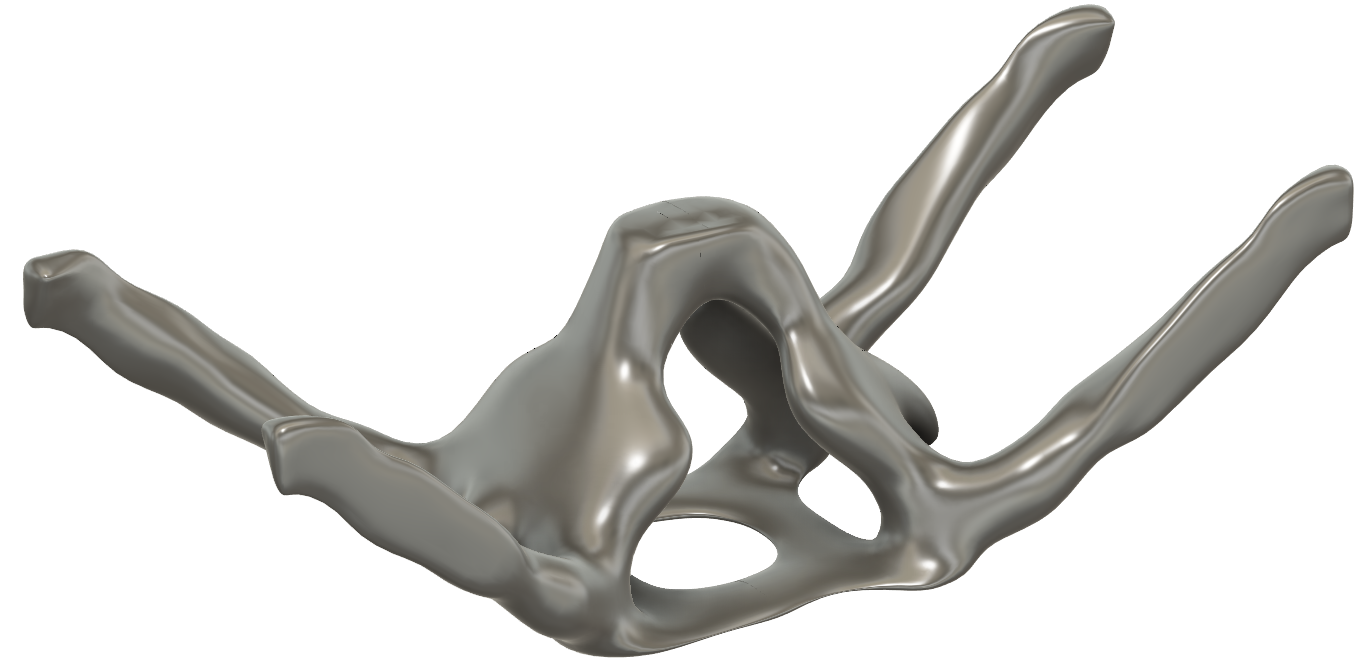}
$c=33$,
$H_{\bar{\alpha}}=0.32$
\vspace{1.0cm}
\vfill
\centering
\includegraphics[width=\textwidth]{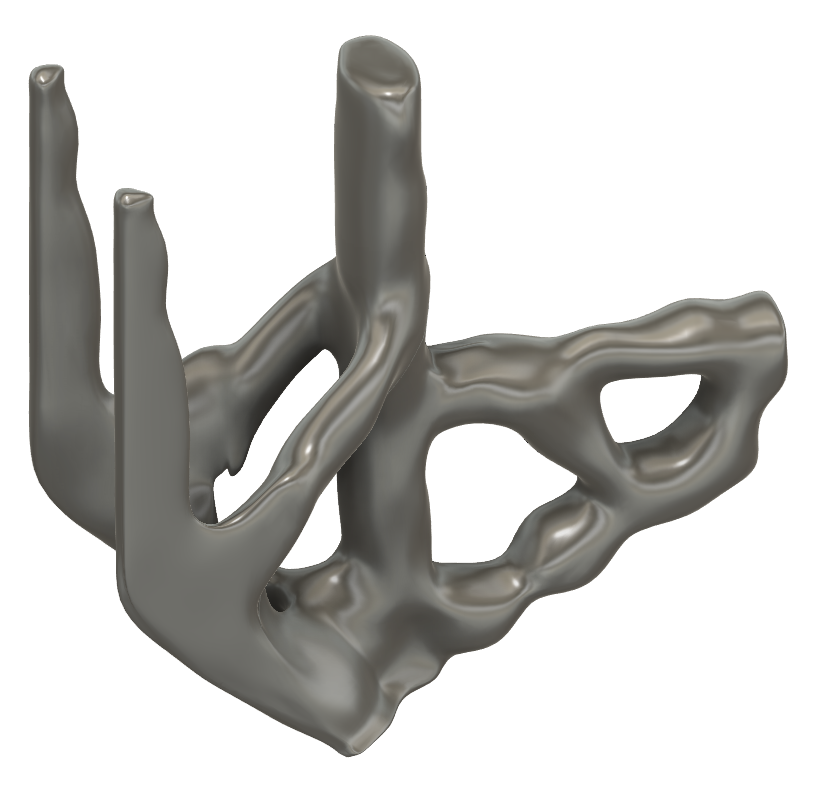}
$c=52$,
$H_{\bar{\alpha}}=0.026$
\vspace{0.2cm}
\caption{}
\end{subfigure}
\qquad
\begin{subfigure}[t]{0.15\textwidth}
\centering
\includegraphics[width=0.8\textwidth]{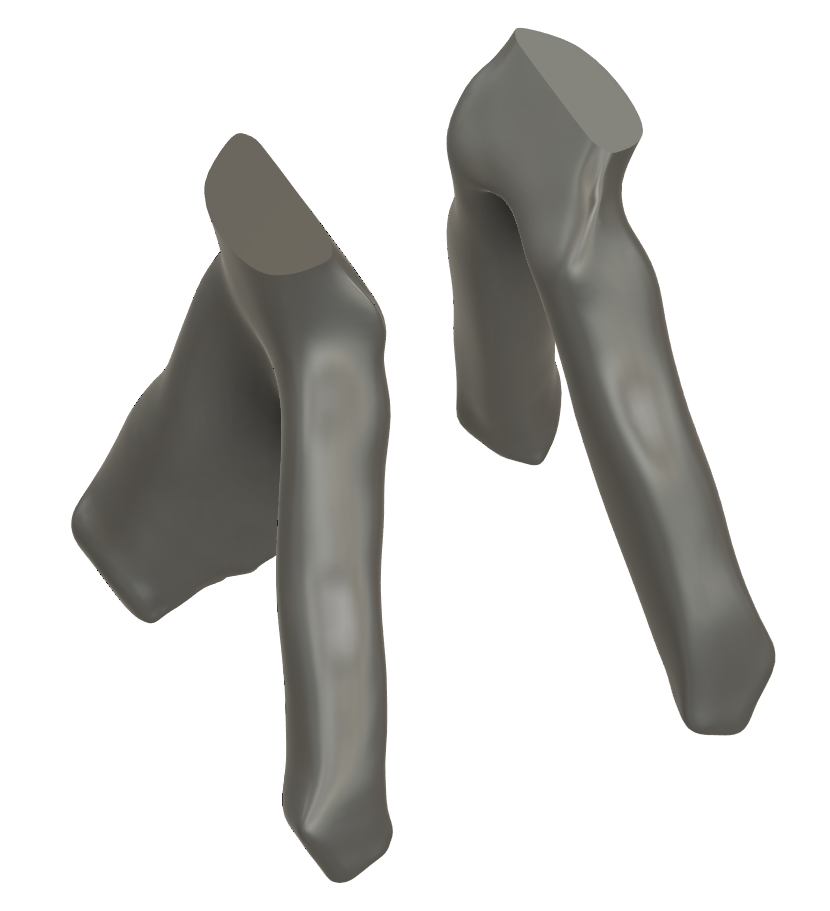}
\vspace{1.2cm}
\vfill
\centering
\includegraphics[width=\textwidth]{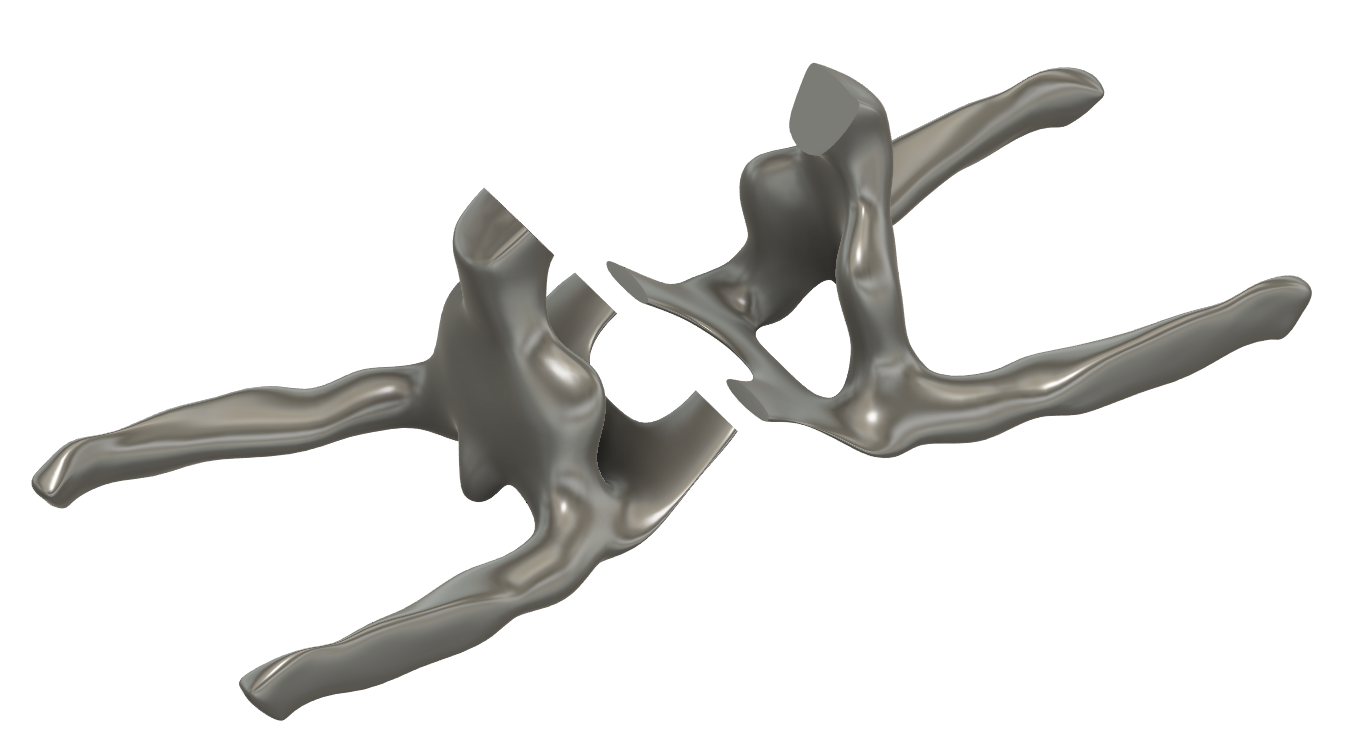}
\vspace{1.2cm}
\vfill
\centering
\includegraphics[width=\textwidth]{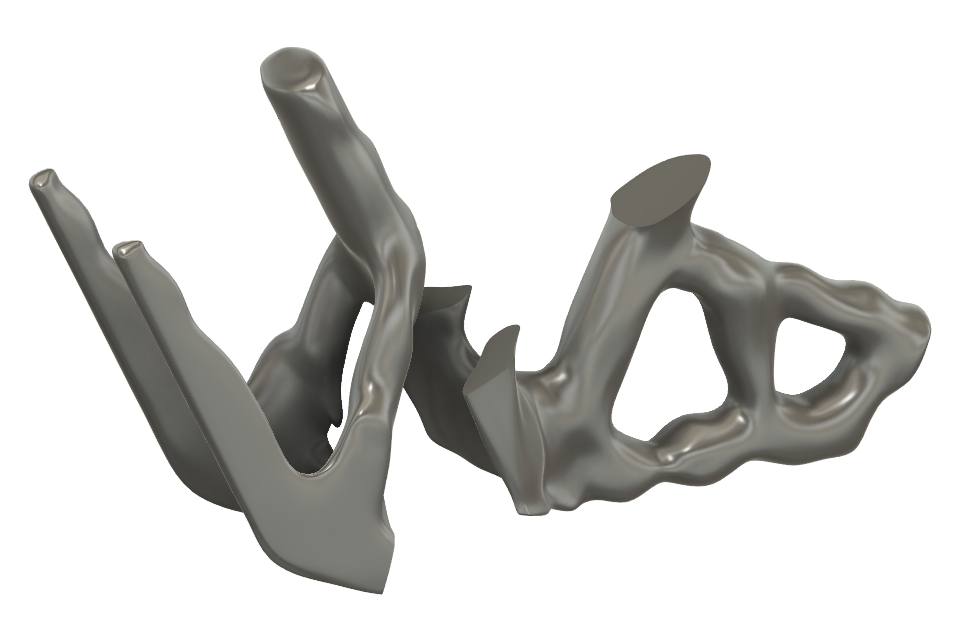}
\vspace{1.4cm}
\caption{}
\end{subfigure}

\centering
\caption{Additional concurrent optimization examples with segmentation and rotation angle applied. (a) Problem boundary condition. (b) Topology optimization only. (c) Concurrent topology, segmentation, and print angle optimization. (d) Segmentation with rotation applied. }

\label{fig:more_demo}
\end{figure*}
\end{document}